\documentclass{article} %
\usepackage{iclr2027_conference,times}

\usepackage{amsmath,amsfonts,bm}

\def\eqref#1{equation~\ref{#1}}

\def\1{\bm{1}}

\DeclareMathAlphabet{\mathsfit}{\encodingdefault}{\sfdefault}{m}{sl}
\SetMathAlphabet{\mathsfit}{bold}{\encodingdefault}{\sfdefault}{bx}{n}

\usepackage{url}            %
\usepackage{booktabs}       %
\usepackage{amsfonts}       %
\usepackage{nicefrac}       %
\usepackage{microtype}      %
\usepackage{xcolor}         %
\usepackage[inline]{enumitem}
\usepackage{amsthm}
\usepackage{bbm}
\usepackage{amsmath}
\usepackage{mathtools}

\usepackage{commands}

\usepackage{lineno}
\usepackage{hyperref}   %

\usepackage{cleveref}
\crefname{section}{\S}{\S\S} %
\Crefname{section}{\S}{\S\S} %
\crefname{table}{Table}{Tables}
\crefname{figure}{Figure}{Figures}
\crefname{algorithm}{Algorithm}{}
\crefname{equation}{Eq.}{Eqs.}
\crefname{appendix}{App.}{App.}
\crefname{line}{Line}{Lines}
\crefformat{section}{\S#2#1#3}
\crefname{definition}{Def.}{Definitions}
\crefname{lemma}{Lemma}{Lemmas}
\crefname{proposition}{Prop.}{Propositions}
\crefname{assumption}{Assumption}{Assumptions}
\crefname{theorem}{Theorem}{Thms.}
\crefname{corollary}{Corollary}{}

\newtheorem{theorem}{Theorem}
\newtheorem{definition}{Definition}
\newtheorem{proposition}{Proposition}
\newtheorem{lemma}{Lemma}
\newtheorem{assumption}{Assumption}
\newtheorem{corollary}{Corollary}

\usepackage[disable]{todonotes}
\newcommand*\iftodonotes{\if@todonotes@disabled\expandafter\@secondoftwo\else\expandafter\@firstoftwo\fi}  %
\makeatother

\definecolor{darkblue}{rgb}{0, 0, 0.5}
\hypersetup{colorlinks=true, citecolor=darkblue, linkcolor=darkblue, urlcolor=darkblue}

\usepackage{thm-restate}
\usepackage{enumitem}

\newenvironment{proofsketch}{\begin{proof}[Proof sketch]}{\end{proof}}

\usepackage{tikz}
\usetikzlibrary{automata, positioning, arrows}
\usepackage{subcaption}

\newtheorem{example}{Example}

\usepackage{placeins}

\title{On the Sample Complexity of Language Generation in the Limit with Noise, Prompts, and Feedback}

\author{Alexandra Butoi \\
  ETH Zürich\\
  \texttt{\href{mailto:abutoi@ethz.ch}{alexandra.butoi@inf.ethz.ch}}
  \And
  Ryan Cotterell \\
  ETH Zürich \\
  \texttt{\href{mailto:rcotterell@ethz.ch}{ryan.cotterell@inf.ethz.ch}}
}

\begin{document}

\maketitle

\begin{abstract}
Language generation in the limit is a learning paradigm in which a learner observes an infinite stream of strings from an unknown target language, chosen from a countable collection, and must eventually produce strings not yet seen. This task is always tractable, i.e., for any countable collection, there exists a learner that eventually succeeds. A natural follow-up question is: how many
examples does the learner need before it can do so? Previous work established upper and lower bounds for regular, locally threshold testable (LTT), and context-free languages. However, the sample complexity has not been studied under the extensions of the framework that model noisy input, conditioning on a prefix, and feedback. In this paper, we analyze the sample complexity of language generation in each of these settings. First, we study the sample complexity for partially-ordered (PO) languages, a class not previously analyzed in this framework, and we establish exponential upper and lower bounds. Next, we show that for context-free languages, the sample bounds remain uncomputable in every setting. Finally, we show that for regular, LTT, and PO languages, finite noise increases the sample complexity by only a linear additive term, prompting leaves the bounds unchanged up to a constant factor in the exponent, and finite feedback can sometimes reduce them exponentially.\looseness=-1
\end{abstract}

\section{Introduction}

Language identification in the limit is an influential learning paradigm \citep{gold1967language, ANGLUIN1980}. In the classic Gold--Angluin model, an adversary enumerates an infinite stream of strings from a target language $\theTrueLanguage$, chosen from a countable collection of languages $\theLanguageCollection$. At each time step, the learner observes the strings enumerated so far and outputs a hypothesis from $\theLanguageCollection$ as its guess for the target language; it succeeds if there exists some finite timestep $\theTimestep$ after which its hypotheses stabilize to the correct language $\theTrueLanguage$. Identification in the limit is impossible in general, and remains so even for relatively simple language classes such as the regular languages \citep{ANGLUIN1980}. Recently, \cite{kleinberg-2024-language-generation} introduced the paradigm of language \emph{generation} in the limit, inspired by modern language models. Rather than identifying the target language $\theTrueLanguage$, the learner must produce, after some finite timestep $\theTimestep$, a previously unseen string from $\theTrueLanguage$. The authors proved that this problem is significantly easier than identification in the limit, showing that it is tractable for every countable collection $\theLanguageCollection$.\looseness=-1

This tractability result sparked significant follow-up work.
\citet{raman-generation-plmr} introduced \emph{uniform} generation in the limit, a
stricter variant requiring the learner to succeed after seeing a number of
strings that is uniform across all languages and enumerations, and showed that
this stronger requirement makes the problem intractable for some countable
collections.
Subsequent work has extended the framework along several axes: prompted
generation \citep{kleinberg-2024-language-generation,raman-generation-plmr}, noisy input
\citep{raman2025generation,mehrotra2025languagegenerationinfinitecontamination, Bai_2026, li2026characterizing}, omissions
\citep{mehrotra2025languagegenerationinfinitecontamination, Bai_2026}, and feedback
\citep{charikar2024exploringfacetslanguagegeneration,Bai_2026}, some within the uniform generation
setting. Other work compared language generation in the limit with established learning
theory frameworks such as PAC or online learning \citep{raman-generation-plmr}, studied language generation in a statistical setting \citep{kalavasis2025limits}, or measured the learner's success by the total
number of mistakes rather than by the time of its last mistake
\citep{kleinberg2026mistakeboundedlanguagegeneration,
jimenez2026polynomialtimemistakeboundedlanguagegeneration,
li2026computationalhardnessmistakeboundedlanguage}; we discuss
these in \cref{sec:related-work}.\looseness=-1

A natural and important question is: how many examples does the learner need
before it can generate correctly? This question is most meaningful in the
uniform generation setting, where the bound must hold irrespective of the target
language or the enumeration. \citet{arenas2026languagegenerationcomplexitybarriers} answered this question
for finite collections, establishing bounds for several language classes: double
exponential for regular languages, exponential for locally threshold testable
(LTT) languages, and uncomputable for context-free languages.
While these classes provide useful theoretical insights, their bounds carry large exponents, and no class is known to do better.
Moreover, even for the language classes studied by \citet{arenas2026languagegenerationcomplexitybarriers}, the
sample complexity under noise, prompting, and feedback is unknown.\looseness=-1 

We first analyze the sample complexity of generation for what we call partially-ordered (PO) languages, a class not previously studied in this framework. By exploiting the structure of the automaton we establish single-exponential upper and lower bounds with a smaller exponent than for LTT languages.
Next, we study the sample complexity in each extension of the framework. We begin by recasting prompted generation and generation with feedback such that they align closer to how language models are used in practice. Where
\citet{raman-generation-plmr} treat a prompt as a label attached to an example, we
follow \citet{kleinberg-2024-language-generation} and treat it as a prefix, with the learner
producing a continuation. \Citet{charikar2024exploringfacetslanguagegeneration} allow one feedback query
per timestep; instead, we follow \citet{Bai_2026} in fixing a total query budget.
We then define uniform generation in each of these settings, characterize which
collections admit uniform generation, and show that uniform generation is always
possible for finite collections. This is a necessary stepping stone without which the
sample complexity bounds cannot be derived.\looseness=-1

We then analyze the sample complexity across all settings and language classes (see \cref{tab:bounds} for a summary of the upper bounds). For context-free languages, the sample complexity remains uncomputable regardless of the setting. For regular, LTT, and PO languages, the picture is more nuanced: finite noise increases the sample complexity by only a linear additive term, prompting leaves the bounds unchanged for regular and PO languages, and for LTT languages changes them only by a constant factor in the exponent, independently of the prompt, and finite feedback can sometimes reduce them exponentially. We complement the upper bounds with lower bounds that match some, but not all, of the terms, and we leave open whether the remaining gaps can be closed.\looseness=-1

Our contributions can be summarized as follows:
\begin{enumerate*}[label=(\roman*)]
\item we isolate the arguments of \citet{arenas2026languagegenerationcomplexitybarriers} into a recipe for deriving sample complexity bounds (\cref{sec:recipe}),
\item we give single-exponential upper and lower bounds for PO languages (\cref{sec:po});
\item we reformulate uniform prompted generation and generation with feedback, and we give definitions of uniform
generatability in each setting, closure dimensions characterizing which
collections admit it, and proofs that finite collections always do (\cref{sec:prompted},\cref{sec:feedback}),
\item we give bounds for sample complexity of generation with finite noise, prompts, and finite feedback for regular, LTT, and PO languages, and
\item we show that the bounds for context-free languages are uncomputable in each setting.
\end{enumerate*}

\section{Preliminaries}\label{sec:flt}

Prior work has established sample complexity bounds for generation in the limit for several well-known classes and subclasses of formal languages \citep{arenas2026languagegenerationcomplexitybarriers}. We define these here.\looseness=-1

Let $\theAlphabet$ be a finite \defn{alphabet} of symbols.
A \defn{string} over $\theAlphabet$ is a sequence of symbols $\theString=\theSymbol_1 \theSymbol_2 \cdots$, where $\theSymbol_i\in\theAlphabet$ for all $i$.
For a string $\theString=\theSymbol_1 \theSymbol_2 \cdots \theSymbol_k$, we use $|\theString|$ to denote its length $k$.
Let $\varepsilon$ denote the \defn{empty string}. A \defn{formal language} $\theLanguage\subseteq\theAlphabet^\ast$ over the alphabet $\theAlphabet$ is a subset of the \defn{Kleene closure} $\theAlphabet^\ast$, i.e., the set of all finite strings over the alphabet $\theAlphabet$. A central quantity in our analysis is the intersection of formal languages. For two languages $\theLanguage_1$ and $\theLanguage_2$, their \defn{intersection} is $\theLanguage_1\cap\theLanguage_2=\{\theString\colon\theString\in\theLanguage_1\text{ and }\theString\in\theLanguage_2\}$.
\looseness=-1

\subsection{Regular languages}
Regular languages are among the simplest and most extensively studied language class, characterized equivalently by finite automata and regular expressions.

\begin{definition}
    A \defn{finite automaton} (FA) is a tuple $\theAutomaton=\theFSATuple$, where $\theAlphabet$ is a finite \defn{alphabet} of symbols, $\theStates$ is a finite set of \defn{states}, $\theInitStates\subseteq \theStates$ is a set of \defn{initial states}, $\theFinalStates\subseteq \theStates$ is a set of \defn{final states}, and $\theTransitions\subseteq \theStates \times \theAlphabet \times \theStates$ is a finite set of \defn{transitions}.
\end{definition}

\begin{definition}
    An FA is called \defn{deterministic FA} (DFA) if $\theInitStates$ consists of at most one initial state $\theInitState$,
    and, for each pair $(\theState,\theSymbol)\in\theStates\times\theAlphabet$, there exists at most one state $\theState'$ such that $(\theState,\theSymbol,
    \theState')\in \theTransitions$; otherwise, it is called \defn{non-deterministic FA} (NFA).\footnote{Generally, NFAs can have transitions of the form $(\theState,\varepsilon,\theState')$ that do not scan any symbol. Such transitions can be easily eliminated without increasing the number of states.}
\end{definition}

A \defn{path} in an FA is a sequence $\thePath=(\theState_0,\theSymbol_1,\theState_1),(\theState_1,\theSymbol_2,\theState_2),\ldots,(\theState_{n-1},\theSymbol_n,\theState_n)$, where $(\theState_{i-1},\theSymbol_i,\theState_i)\in\theTransitions$ for all $i\in[1,n]$. If $\theState_0\in\theInitStates$ and $\theState_n\in\theFinalStates$, the path $\thePath$ is called \defn{accepting}. The path $\thePath$ is called \defn{simple} if all its states are distinct; otherwise, it has a \defn{cycle}.
FAs recognizing finite languages do not have cycles.
The \defn{length} of $\thePath$ is the number of transitions it contains. 
If $\theString=\theSymbol_1\cdots\theSymbol_n\in\theAlphabet^\ast$ for such an accepting path, then $\theAutomaton$ \defn{recognizes} the string $\theString$. 
In an FA without cycles, the length of the longest simple path coincides with the length of the longest recognized string.
The \defn{language} $L(\theAutomaton)$ of an FA $\theAutomaton$ is the set of all strings it recognizes. A language is \defn{regular} if there is some FA recognizing it.
Any NFA can be converted into an equivalent DFA, but this transformation may incur an exponential blowup in the number of states \citep{hopcroft-2006-introduction}.
The intersection of two regular languages is also regular, and it can be computed via the \defn{product construction} \citep{hopcroft-2006-introduction}, which constructs an NFA that simulates both NFAs in parallel. The resulting product NFA has at most as many states as the product of the sizes of the two original NFAs.\looseness=-1

\begin{table}[t]
\centering
\caption{Upper bounds of sample complexity of uniform generation in the limit for regular, locally threshold testable (\textbf{LTT}), partially-ordered (\textbf{PO}), and context-free languages across the base setting of \cite{kleinberg-2024-language-generation}, generation with noise, prompted generation, and generation with feedback. We sometimes give the bounds in terms of the base bound $B$.
We use $s$ to denote the largest size of the representation of each language, $k$ the noise
level, and $f < |\theLanguageCollection|-1$ the feedback budget. The notation $x \mapsto y$ denotes that $x$ gets replaced by $y$ in the base bound.\looseness=-1}
\label{tab:bounds}
\small
\begin{tabular}{lllll}
\toprule
 \textbf{Language class}& \textbf{Base ($B$)} & \textbf{Noise $k$} & \textbf{Prompt} & \textbf{Feedback $f$} \\
\midrule
\textbf{Regular} & $|\Sigma|^{s^{|\mathcal{C}|}}$
        & $B + k|\mathcal{C}|$
        & $B$
        & $|\Sigma|^{s^{(|\mathcal{C}|-f)}}$ \\
\textbf{LTT}     & $|\Sigma|^{3r(r^2+1)^2}$, $r=s|\mathcal{C}|$
        & $B + k|\mathcal{C}|$
        & $r \mapsto 2s|\mathcal{C}|$
        & $r \mapsto s(|\mathcal{C}|{-}f)$ \\
\textbf{PO}      & $|\Sigma|^{|\mathcal{C}|(s-1)+1}$
        & $B + k|\mathcal{C}|$
        & $B$
        & $|\Sigma|^{(|\mathcal{C}|-f)(s-1)+1}$ \\
\textbf{Context-free}     & uncomputable
        & uncomputable
        & uncomputable
        & uncomputable \\
\bottomrule
\end{tabular}
\end{table}

\subsection{Locally threshold testable languages} 
Locally threshold testable (LTT) languages are a well-studied subclass of the regular languages. They admit several equivalent characterizations, including one in terms of profile sequences.

\begin{definition}
    A \defn{profile} over an alphabet $\theAlphabet$ is a tuple $\theProfile=(Pr, Su, In)$, where $Pr$ and $Su$ are a prefix and a suffix, respectively, and $In$ is a finite set of constraints of the form $(\theString,\theta t)$, where $\theString\in\theAlphabet^\ast$, $\theta \in \{\leq, <, =, \geq, >\}$, and $t\in\N$, given in unary notation.\footnote{This means that $t$ is written as a string $\theSymbol^t$ for some symbol $\theSymbol\in\theAlphabet$.}
\end{definition}

The language $L(\theProfile)$ defined by the profile $\theProfile$ is the set of all strings $\theString\in\theAlphabet^\ast$ such that $Pr$ is a prefix of $\theString$, $Su$ is a suffix of $\theString$, and for every $(y, \theta t)\in In$, the number of occurrences of the substring $y$ satisfies the constraint $\theta t$. 
A language $L$ is \defn{locally threshold testable} (LTT) if there exists a finite sequence of profiles $(\theProfile_1,\ldots, \theProfile_n)$ such that $L=\bigcup_{i=1}^n L(\theProfile_i)$.
The \defn{size} of a profile is the length of its representation as a string, i.e., $P$ written as $(\mathit{Pr}, \mathit{Su}, \{(x_1, \theta_1 t_1), \ldots, (x_m, \theta_m t_m)\})$ with each $t_i$ in unary, counting all parentheses, commas, braces and comparison symbols. In particular, $|P| \geq 1 + |\mathit{Pr}| + |\mathit{Su}| + \sum_{i=1}^{m} (|x_i| + t_i + 1)$.
Every string in the language of a profile of size $s$ that defines a finite language has length less than $3s(s^2+1)^2$ \citep[Lemma 5.2]{arenas2026languagegenerationcomplexitybarriers}.\looseness=-1

\subsection{Context-free languages}
Context-free languages are another important language class, capturing hierarchical structure that simpler classes cannot express. Context-free grammars are the canonical formalism for defining them; we define them formally and give an example in \cref{sec:cf}.
Computing the size of the intersection of two context-free languages is undecidable, and \citet{arenas2026languagegenerationcomplexitybarriers} show that this carries over to the sample complexity via a reduction from the halting problem: a computable bound would allow deciding whether a Turing machine halts, contradicting undecidability \citep{hopcroft-2006-introduction}. We refer the reader to \cref{sec:turing-machines} for background on Turing machines and the halting problem. We build on this reduction to establish uncomputability in all of our settings.\looseness=-1

\section{Uniform language generation in the limit}\label{sec:language-generation}

\paragraph{The setup.}
Let $\theLanguageCollection=\{\theLanguage_1, \theLanguage_2, \ldots\}$ be a \defn{countable collection} of languages over a \defn{countable domain} $\theDomain$, i.e., each $\theLanguage_i$ is a subset of $\theDomain$ ($\theLanguage_i \subseteq \theDomain$).
Throughout the rest of the paper, we assume the domain is $\theAlphabet^\ast$.
All languages $\theLanguage_i$ are \emph{infinite}. An adversary and a learner play the following game. The \defn{adversary} chooses a \defn{target} language $\theTrueLanguage=\theLanguage_\theTrueLanguageIndex\in\theLanguageCollection$ and enumerates strings $\theCurrentString$ from it, one by one, over a sequence of steps $\theTimestep=1,2,3,\ldots$.
At timestep $\theTimestep$, the \defn{enumeration} of strings shown so far is denoted by $\theCurrentStringSet$. 
Every string in $\theTrueLanguage$ must appear at some timestep $\theTimestep$, but repetitions are allowed. 
In \defn{noisy generation}, the enumeration may contain noisy examples $\theString\notin\theTrueLanguage$, but it still needs to cover all of $\theTrueLanguage$.
We sometimes use $\theCurrentStringSet$ to denote the set of distinct strings in the enumeration $\theCurrentStringSet$. 
In each step, the \defn{learner} $\learner$ observes the enumeration $\theCurrentStringSet$ and must output one \emph{new} string $\theCurrentGeneratedString=\learner(\theCurrentStringSet)$ from $\theTrueLanguage$.
In \defn{generation with feedback}, the learner is allowed to ask whether a string $\theString$ belongs to $\theTrueLanguage$ and the adversary must answer truthfully.
In \defn{prompted generation}, the adversary additionally shows a prompt $\thePrompt\in\thePromptDomain$ and the learner must output a new continuation $\theString$ such that $\thePrompt\cdot\theString\in\theTrueLanguage$.
Let $\theStringSet=\bigcup_{\theTimestep=1}^\infty \theCurrentStringSet$. A learner \defn{generates in the limit} if for any target language $\theTrueLanguage$ in $\theLanguageCollection$ and any enumeration $\theStringSet$, there is a timestep $\theTimestepStar$ such that for all steps $\theTimestep \geq \theTimestepStar$, the algorithm's guess is correct \citep{kleinberg-2024-language-generation}. \looseness=-1

Throughout, we assume the following: (i) the target language $\theTrueLanguage$ belongs to the collection $\theLanguageCollection$, (ii) membership in a language is binary, (iii) every language in $\theLanguageCollection$ is infinite, (iv) $\theLanguageCollection$ is finite, and (v) feedback is truthful. Assumption (iv), although restrictive, is essential for deriving the bounds.\looseness=-1

\paragraph{Uniform generation in the limit.}
The timestep $\theTimestep^\ast$ can depend both on the target language $\theTrueLanguage$, as well as the enumeration $\theStringSet$. In follow-up work, \citet{raman-generation-plmr} introduced  \defn{uniform generation}, which requires this timestep to be \emph{uniform} across all languages and enumerations.\looseness=-1

\begin{definition}[\citealp{raman-generation-plmr}]
A collection $\theLanguageCollection$ is \defn{uniformly generatable in the limit} if there is an algorithm $\learner$ and a sample threshold $\distinctNumExamples$ such that for any language $\theTrueLanguage\in\theLanguageCollection$, any enumeration of strings $\theString_1,\theString_2,\ldots \in \theTrueLanguage$, if there exists a timestep $\theTimestep^\ast$ such that $|\theStringSet_{\theTimestep^\ast}|=\distinctNumExamples$, then for all $\theTimestep\geq\theTimestep^\ast$ it holds that $\theGeneratedString_\theTimestep\in\theTrueLanguage\setminus\theStringSet_\theTimestep$.\looseness=-1
\end{definition}

Unlike generation in the limit, \emph{uniform} generation in the limit is not always possible: \citet{raman-generation-plmr} showed which collections admit it using the following characterization.\looseness=-1

\begin{definition}[\citealp{raman-generation-plmr}]

    For any collection of languages $\theLanguageCollection$ and enumeration $\theStringSet_n=\theString_1, \theString_2, \ldots, \theString_n$, let the set $\theLanguageCollection (\theStringSet_n) = \{\theLanguage\in\theLanguageCollection \colon \theStringSet_n \subseteq \theLanguage\}$
    be the set of \defn{consistent languages} with the enumeration $\theStringSet_n$. Then, we define the \defn{closure} $\left<\cdot\right>_\theLanguageCollection$ as
    \begin{equation}
        \left<\theStringSet_n\right>_\theLanguageCollection = 
        \begin{cases}\bigcap_{\theLanguage\in\theLanguageCollection (\theStringSet_n)} \theLanguage, & \text{if } \theLanguageCollection (\theStringSet_n) \neq \emptyset \\
        \bot, &\text{if } \theLanguageCollection (\theStringSet_n) = \emptyset.
        \end{cases}
    \end{equation}
\end{definition}
Intuitively, the closure is the intersection of languages that are consistent with the examples seen so far.\looseness=-1

\begin{definition}[\citealp{raman-generation-plmr}]
    The \defn{closure dimension} $\closureDim(\theLanguageCollection)$ of $\theLanguageCollection$ is the supremum over $d\in\mathbb{N}$ for which there exist distinct strings $\theString_1,\theString_2, \ldots, \theString_d\in\theDomain$ such that $\left<\{\theString_1,\theString_2,\ldots,\theString_d\}\right>_\theLanguageCollection\neq\bot$ and $|\left<\{\theString_1,\theString_2,\ldots,\theString_d\}\right>_\theLanguageCollection|< \infty$. If it is not true for $d=1$, we set $\closureDim(\theLanguageCollection)=0$, and if the set of such $d$ is unbounded, we set $\closureDim(\theLanguageCollection)=\infty$.
\end{definition}

The sample complexity of uniform generation is only well-defined for collections that are uniformly generatable. \citet{raman-generation-plmr} gave the following characterization.\looseness=-1

\begin{theorem}[\citealp{raman-generation-plmr}]
    If $\closureDim(\theLanguageCollection)<\infty$, then $\theLanguageCollection$ is uniformly generatable.
\end{theorem}
A finite closure dimension guarantees that once the learner has seen enough distinct examples, the intersection of all consistent languages is always infinite, so the learner can never be trapped with only finitely many valid strings to output. For finite collections, the closure dimension is always finite since there are only finitely many possible intersections \citep{raman-generation-plmr}. We say that a collection is \defn{$\distinctNumExamples$-generatable} if uniform generation succeeds after observing $\distinctNumExamples$ distinct examples.

\section{A recipe for deriving sample complexity bounds}
\label{sec:recipe}

All the sample complexity bounds for uniform generation rest on the following
characterization, which reduces $\distinctNumExamples$-generatability to a question about the structure
of the language class.

\begin{proposition}[\citealp{arenas2026languagegenerationcomplexitybarriers}]\label{prop:arenas-prop-two-three}
    A collection $\theLanguageCollection$ of infinite languages is \emph{not} $\distinctNumExamples$-generatable if and only if there exists a nonempty subset $\theLanguageCollSubset\subseteq \theLanguageCollection$ such that $\bigcap_{\theLanguage\in\theLanguageCollSubset}\theLanguage$ is finite and has size at least $\distinctNumExamples$.
\end{proposition}
It therefore suffices to bound the size of the largest finite intersection of
languages in the collection. Every bound in this paper and in
\citet{arenas2026languagegenerationcomplexitybarriers} uses
the following four steps.\looseness=-1

\begin{enumerate}[leftmargin=2em,itemsep=1pt,topsep=2pt,parsep=0pt, label=(\roman*)]
  \item \label{item:step-one} \textbf{Close the representation under intersection.} Fix a
    representation (e.g., automaton, profile, etc.) for the class and a construction mapping representations of
    $\theLanguage_1, \dots, \theLanguage_n$ to a representation of
    $\bigcap_i \theLanguage_i$, together with a bound $\sigma(s, n)$ on the size
    of the result when each of the $n$ inputs has size at most $s$. For instance, for automata this is
    the product construction, with $\sigma(s,n) = s^{n}$ states.\looseness=-1
  \item \label{item:step-two} \textbf{Bound the longest string.} Show that, if
    the intersection is finite, then every string it contains has length at most
    $\ell$, depending on $\sigma(s,n)$. Finiteness is what makes this possible. For automata,
    it forbids cycles, so the longest accepted string is
    bounded by the longest simple path.
  \item \textbf{Count strings.} A language over an alphabet $\theAlphabet$ with $|\theAlphabet|\ge 2$, whose strings have length at most $\ell$,
    contains at most $\sum_{i=0}^{\ell}|\Sigma|^{i} =
    \frac{|\Sigma|^{\ell+1}-1}{|\Sigma|-1} < |\Sigma|^{\ell+1}$ strings. \label{item:recipe-step-three}
  \item \label{item:step-four} \textbf{Apply \cref{prop:arenas-prop-two-three}.} The collection
    is $\distinctNumExamples$-generatable for any $\distinctNumExamples$ exceeding the bound from \cref{item:recipe-step-three}.
\end{enumerate}

Notice that only \cref{item:step-two} is class-specific; \cref{item:step-one}, \cref{item:recipe-step-three} and \cref{item:step-four} are generic.
Moreover, \cref{item:step-one} only applies to representations that are closed under intersection.
For a collection $\theLanguageCollection$, we write $B(\theLanguageCollection) = |\theAlphabet|^{\ell+1}$,
where $\ell$ is the length bound from \cref{item:step-two}, so that every finite intersection of
a non-empty subcollection of $\theLanguageCollection$ has size less than
$B(\theLanguageCollection)$. By
\cref{prop:arenas-prop-two-three}, $\theLanguageCollection$ is
$B(\theLanguageCollection)$-generatable.

\paragraph{The role of the right representation.}
Because $\ell$ carries the entire bound, the choice of representation matters
more than its compactness. Regular languages are far more succinctly described
in monadic second-order (MSO) logic than by automata, but the longest string in a
finite intersection of MSO-definable languages is an exponential tower
\citep{hella2022}, so the resulting $\ell$ is vastly worse than the
automaton-based one. 
For PO languages, we do not change the representation, but exploit
additional structure inside it to bound the longest string. Therefore, the right representation is
the one that most tightly bounds $\ell$, and it depends on the language class.\looseness=-1

We briefly summarize the constructions of \cite{arenas2026languagegenerationcomplexitybarriers} for regular, LTT and context-free languages:
\begin{itemize}[leftmargin=2em,itemsep=1pt,topsep=2pt,parsep=0pt]
\item \textbf{Regular languages:} The product of $|\mathcal{C}|$ NFAs with at most $s$ states each has up to
$s^{|\mathcal{C}|}$ states. If the intersection is finite,
no accepting path can traverse a cycle, so every accepting path is simple and visits each
state at most once. The longest accepted string is then bounded by the number of
states, i.e., $\ell = s^{|\mathcal{C}|} - 1$. Counting strings of up to this length gives a double exponential bound.
\item \textbf{LTT languages:} Intersecting profiles amounts to
collecting their prefix, suffix and counting constraints together, so
$|\mathcal{C}|$ profiles of size at most $s$ yield a profile of size at most
$s|\mathcal{C}|$. Finiteness then
forces the counting constraints to cap the number of occurrences of every
substring, which caps the length of any string in the language: a profile of
size $r$ defining a finite language admits no string longer than $3r(r^2+1)^2$
\citep[Lemma 5.2]{arenas2026languagegenerationcomplexitybarriers}. With $r = s|\mathcal{C}|$, the length
bound $\ell$ is polynomial in $s$ and $|\mathcal{C}|$, and the sample complexity
drops to a single exponential.
\item \textbf{Context-free languages:} The size of the intersection of two context-free languages is not computable from the grammars: one can
construct two grammars whose intersection has size exactly $2^t$, where $t$ is
the number of steps a given Turing machine takes to halt
\citep[Lemma 3.2]{arenas2026languagegenerationcomplexitybarriers}. A computable length bound would yield a
computable bound on $t$ and decide the halting problem.
\end{itemize}

\section{Sample complexity of generation for partially-ordered languages}\label{sec:po}

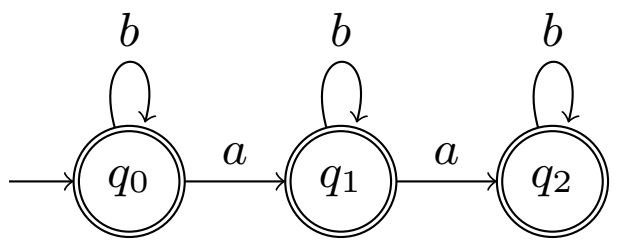
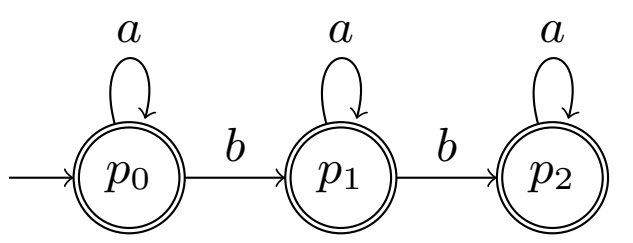
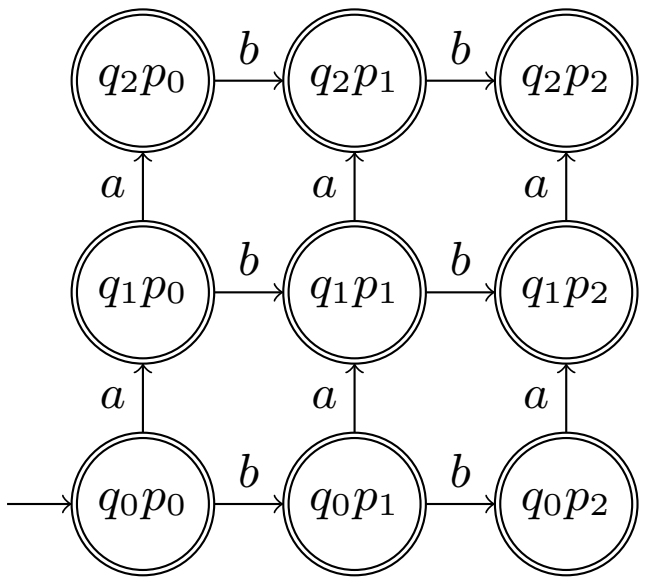
\begin{figure}[tp]
    \centering
    \footnotesize
    \begin{minipage}[b]{0.35\textwidth}
        \begin{subfigure}[b]{\textwidth}
        \centering
        \begin{tikzpicture}[
            ->, auto, node distance=1.4cm,
            initial text={},
            every accepting/.style={double}
        ]
            \node[state, initial, accepting, minimum size=0.7cm, font=\small] (q0) {$q_0$};
            \node[state, accepting, minimum size=0.7cm, font=\small] (q1) [right of=q0] {$q_1$};
            \node[state, accepting, minimum size=0.7cm, font=\small] (q2) [right of=q1] {$q_2$};
            \path (q0) edge [] node[font=\small] {$a$} (q1)
            (q1) edge [] node[font=\small] {$a$} (q2)
                  (q0) edge [loop above] node[font=\small] {$b$} (q0)
                  (q1) edge [loop above] node[font=\small] {$b$} (q1)
                  (q2) edge [loop above] node[font=\small] {$b$} (q2)
                  ;
        \end{tikzpicture}
        \subcaption{PONFA recognizing language $\theLanguage_1=\{\theString\colon\theString\text{ has at most two }a\text{s}\}$.}
    \end{subfigure}
    \hfill
    \begin{subfigure}[b]{\textwidth}
        \centering
        \begin{tikzpicture}[
            ->, auto, node distance=1.4cm,
            initial text={},
            every accepting/.style={double}
        ]
            \node[state, initial, accepting, minimum size=0.7cm, font=\small] (p0) {$p_0$};
            \node[state, accepting, minimum size=0.7cm, font=\small] (p1) [right of=p0] {$p_1$};
            \node[state, accepting, minimum size=0.7cm, font=\small] (p2) [right of=p1] {$p_2$};
            \path (p0) edge [loop above] node[font=\small] {$a$} (p0)
                  (p0) edge [] node[font=\small] {$b$} (p1)
                  (p1) edge [] node[font=\small] {$b$} (p2)
                  (p1) edge [loop above] node[font=\small] {$a$} (p1)
                  (p2) edge [loop above] node[font=\small] {$a$} (p2);
        \end{tikzpicture}
        \subcaption{PONFA recognizing language $\theLanguage_2=\{\theString\colon\theString\text{ has at most two }b\text{s}\}$.}
    \end{subfigure}
    \end{minipage}
    \hfill
    \begin{minipage}[b]{0.44\textwidth}
        \centering
        \begin{subfigure}[b]{\textwidth}
        \centering
        \begin{tikzpicture}[
            ->, auto, node distance=1.4cm,
            initial text={},
            every accepting/.style={double}
        ]
            \node[state, initial, accepting, minimum size=0.6cm, font=\footnotesize] (q0p0) {$q_0p_0$};
            \node[state, accepting, minimum size=0.6cm, font=\footnotesize] (q1p0) [above of=q0p0] {$q_1p_0$};
            \node[state, accepting, minimum size=0.6cm, font=\footnotesize] (q2p0) [above of=q1p0] {$q_2p_0$};
            \node[state, accepting, minimum size=0.6cm, font=\footnotesize] (q0p1) [right of=q0p0] {$q_0p_1$};
            \node[state, accepting, minimum size=0.6cm, font=\footnotesize] (q0p2) [right of=q0p1] {$q_0p_2$};
            \node[state, accepting, minimum size=0.6cm, font=\footnotesize] (q1p1) [above of=q0p1] {$q_1p_1$};
            \node[state, accepting, minimum size=0.6cm, font=\footnotesize] (q2p1) [above of=q1p1] {$q_2p_1$};
            \node[state, accepting, minimum size=0.6cm, font=\footnotesize] (q1p2) [right of=q1p1] {$q_1p_2$};
            \node[state, accepting, minimum size=0.6cm, font=\footnotesize] (q2p2) [right of=q2p1] {$q_2p_2$};
            \path (q0p0) edge [] node[font=\small] {$b$} (q0p1)
            (q0p1) edge [] node[font=\small] {$b$} (q0p2)
            (q1p0) edge [] node[font=\small] {$b$} (q1p1)
            (q1p1) edge [] node[font=\small] {$b$} (q1p2)
            (q2p0) edge [] node[font=\small] {$b$} (q2p1)
            (q2p1) edge [] node[font=\small] {$b$} (q2p2)
            (q0p0) edge [] node[font=\small] {$a$} (q1p0)
            (q0p1) edge [] node[font=\small] {$a$} (q1p1)
            (q0p2) edge [] node[font=\small] {$a$} (q1p2)
            (q1p0) edge [] node[font=\small] {$a$} (q2p0)
            (q1p1) edge [] node[font=\small] {$a$} (q2p1)
            (q1p2) edge [] node[font=\small] {$a$} (q2p2);
        \end{tikzpicture}
        \subcaption{PONFA recognizing $\theLanguage_1 \cap \theLanguage_2=\{\theString\colon\theString\text{ has at most two }a\text{s and at most two }b\text{s}\}$.}
        \end{subfigure}
    \end{minipage}
    \caption{Two PONFAs and their product PONFA. The longest simple path has length $2\times 2 =4$.}
    \label{fig:ponfa-example}
\end{figure}

\citet{arenas2026languagegenerationcomplexitybarriers} show an exponential lower bound for non-erasing pattern languages, despite this class being identifiable in the limit. They conclude that identifiability alone does not guarantee sample-efficient generation, and leave open under which structural conditions language generation can be done efficiently. Partially-ordered languages, another subclass of the regular languages, are a step in this direction. Their automata have no cycles other than self-loops, which limits the length of the strings in a finite intersection. This reduces the double-exponential bound for general regular languages to a single-exponential one, with the smallest known exponent in this framework.

The \defn{reachability relation} on a set of states $\theStates$ is defined $\theState \leq \theState'$ if there exists a path from $\theState$ to $\theState'$. 

\begin{definition}
    An NFA is a \defn{partially-ordered NFA} (PONFA) if $\leq$ is a partial order.
\end{definition}

We call a language \defn{partially-ordered} (PO) if and only if it can be recognized by a PONFA.\footnote{These languages are also known as the languages at level $3/2$ of the Straubing--Thérien hierarchy \citep{Schwentick-2001-partially}. Here, we call them \emph{partially-ordered} for simplicity, from the automaton characterizing them. The same bounds hold for partially-ordered DFAs, which are also known as languages with $\mathcal{R}$-trivial monoids \citep{BRZOZOWSKI198032}. Crucially, determinizing PONFAs may not preserve the partial order.\looseness=-1}
The intersection of two PO languages is PO (see \cref{fig:ponfa-example} for an illustration). Using \cref{prop:arenas-prop-two-three} and the depth argument of \citet[Theorem~3.3]{Masopust_2021}, we derive the following
upper bound.\looseness=-1

\begin{restatable}{theorem}{TheoremLTBounds}\label{thm:lt-bound}
Let $\theLanguageCollection$ be a finite collection of infinite PO languages, over an alphabet of size $\theAlphabetSize$ such that $\theAlphabetSize\ge 2$, each recognized by a PONFA with at most $s$ states. Then $\theLanguageCollection$ is $\theAlphabetSize^{|\theLanguageCollection|(s-1)+1}$-generatable.
\end{restatable}

\begin{proofsketch}
The product of PONFAs is also partially-ordered, with states $(\theState_1,\ldots,\theState_{|\theLanguageCollection|})\in\theStates_1 \times \cdots \times \theStates_{|\theLanguageCollection|}$, where $\theStates_i$ is the set of states of the $i^\text{th}$ PONFA. Since the intersection is finite, every accepting path is simple.
Because of the partial order, any transition has the form $(\theState_1,\ldots,\theState_{|\theLanguageCollection|})\xrightarrow{\theSymbol}(p_1,\ldots,p_{|\theLanguageCollection|})$, where $\theState_i \le p_i$ for all $i$, and $\theState_i \neq p_i$ for at least one $i$. That means that, the path must, at each step, transition to a strictly higher tuple of states, therefore at least one automaton must transition to a strictly higher state in its own partial order (see \cref{fig:ponfa-example} for an example). 
Since each PONFA has at most $s$ states, each component can increase at most $s-1$ times, so the longest accepted string has length at most $|\theLanguageCollection|(s-1)$. The intersection therefore has size less than $\theAlphabetSize^{|\theLanguageCollection|(s-1)+1}$, and the bound follows from \cref{prop:arenas-prop-two-three}. The full proof is in \cref{sec:app-piecewise-testable}.
\end{proofsketch}

In \cref{sec:app-piecewise-testable}, we construct, for every $s\geq 2$, a collection of two infinite languages over a binary alphabet, each recognized by a PONFA with $s$ states, whose intersection has $\binom{2s}{s}-1=\Theta(4^s/\sqrt{s})$ strings. This gives an exponential lower bound in $s$. For these collections, the upper bound from \cref{thm:lt-bound} is $2^{2s-1}$, so the upper and lower bounds match up to a factor of $\Theta(\sqrt{s})$. Notice that this shows a near-tight dependence on the number of states for a fixed binary alphabet and two languages, but does not establish matching dependence on all parameters.

\section{Sample complexity of generation with finite noise}

Enumerations may inevitably contain errors, as any real-world corpus would. We study the sample complexity of uniform generation when the learner is guaranteed that at most $\noiseLevel$ of the observed examples are noisy. We build on the noisy generation setting of \citet{raman2025generation} and the characterization of uniform generation at a fixed noise level of \citet{li2026characterizing}. \citet{li2026characterizing} characterize which collections admit generation in this setting. Instead, we derive explicit bounds on the number of required examples. For regular, LTT, and PO languages, introducing noise adds only a linear term in $\noiseLevel$ to the sample complexity. For context-free languages, the uncomputability barrier is preserved.\looseness=-1
\begin{definition}[\citealp{raman2025generation}]
\label{def:noisy-closure}
    For any collection of languages $\theLanguageCollection$, a noisy enumeration $\theStringSet_\theNumExamples=\theString_1, \theString_2, \ldots, \theString_\theNumExamples$, and a noise level $\noiseLevel$, let $\theLanguageCollection (\theStringSet_\theNumExamples,\noiseLevel) = \{\theLanguage\in\theLanguageCollection \colon |\theStringSet_\theNumExamples \cap \theLanguage| \geq |\theStringSet_\theNumExamples|-\noiseLevel\}$
    be the set of consistent languages up to $\noiseLevel$ errors. Then we define the \defn{noisy closure} $\left<\cdot, \noiseLevel\right>_\theLanguageCollection$ at noise level $\noiseLevel$ as
    \begin{equation}
        \left<\theStringSet_\theNumExamples,\noiseLevel\right>_\theLanguageCollection = 
        \begin{cases}\bigcap_{\theLanguage\in\theLanguageCollection (\theStringSet_\theNumExamples, \noiseLevel)} \theLanguage, & \text{if } \theLanguageCollection (\theStringSet_\theNumExamples, \noiseLevel) \neq \emptyset \\
        \bot, &\text{if } \theLanguageCollection (\theStringSet_\theNumExamples, \noiseLevel) = \emptyset.
        \end{cases}
    \end{equation}
\end{definition}
A noisy closure at level $\noiseLevel$ is the intersection of all languages consistent with the enumeration up to $\noiseLevel$ errors. We say a collection is $\distinctNumExamples$-generatable at noise level $\noiseLevel$ if uniform generation succeeds after observing $\distinctNumExamples$ distinct examples, at most $\noiseLevel$ of which are noise.
We defer the formal definitions and generatability results to \cref{sec:noisy-gen-bounds-app}.  The following characterization is the finite-noise analogue of Proposition~2.3 of \citet{arenas2026languagegenerationcomplexitybarriers}, and follows easily from the proof of Corollary~3.4 of \citet{raman2025generation}. Here, we state it in the form we need.

\begin{restatable}{proposition}{NoisyUnifGenerationProposition}\label{prop:noisy-unif-gen}
    A finite collection $\theLanguageCollection$ of infinite languages is \emph{not} uniformly $\distinctNumExamples$-generatable at noise level $\noiseLevel\in\mathbb{N}$ if and only if 
    there exists an enumeration $\theCurrentStringSet=\theString_1,\theString_2,\ldots,\theCurrentString$ with $|\{\theString_1,\ldots, \theString_\theTimestep\}|=\distinctNumExamples$ such that $\theLanguageCollection(\theCurrentStringSet,\noiseLevel)$ is non-empty and $\left<\theStringSet_\theTimestep,\noiseLevel\right>_\theLanguageCollection$ is finite and has size at least $\distinctNumExamples-\noiseLevel |\theLanguageCollection|$.
\end{restatable}
We give a proof in \cref{sec:noisy-gen-bounds-app}.
For all language classes, the noisy bounds follow by applying the same intersection argument as in the noiseless setting. By \cref{prop:noisy-unif-gen}, generatability fails only when the noisy closure has size at least $\distinctNumExamples-\noiseLevel|\theLanguageCollection|$, so it suffices to take $\distinctNumExamples$ larger than the noiseless closure bound plus an additive $\noiseLevel|\theLanguageCollection|$ term. We formalize this intuition in the following theorem. 

\begin{restatable}{theorem}{ThmNoiseGeneric}
\label{thm:noise-generic}
Let $\theLanguageCollection$ be a finite collection of infinite languages that is
$B(\theLanguageCollection)$-generatable. Then $\theLanguageCollection$ is
$(B(\theLanguageCollection) + \noiseLevel|\theLanguageCollection|)$-generatable at noise level $\noiseLevel$.
\end{restatable}

\begin{proof}[Proof sketch]
By \cref{prop:noisy-unif-gen}, it suffices to show that whenever
$\theLanguageCollection(\theCurrentStringSet, \noiseLevel)$ is non-empty and $\langle \theCurrentStringSet, \noiseLevel \rangle_{\theLanguageCollection}$ is
finite, its size is below $\distinctNumExamples - \noiseLevel|\theLanguageCollection|$. The noisy closure is the
intersection of the languages in $\theLanguageCollection(\theCurrentStringSet, \noiseLevel)$, a subset of
$\theLanguageCollection$, so the noiseless bound applies and
$|\langle \theCurrentStringSet, \noiseLevel \rangle_{\theLanguageCollection}| < B(\theLanguageCollection)$. Taking
$\distinctNumExamples \geq B(\theLanguageCollection) + \noiseLevel|\theLanguageCollection|$ makes the condition of
\Cref{prop:noisy-unif-gen} unsatisfiable. We give the full proof in
\cref{sec:noisy-gen-bounds-app}.
\end{proof}

Noise is therefore additive and linear in $\noiseLevel$ and the size of the collection. Instantiating
$B(\theLanguageCollection)$ with the bounds of \cite{arenas2026languagegenerationcomplexitybarriers}, and our bound for PO languages from \cref{sec:po} gives the following:
$\theAlphabetSize^{s^{|\theLanguageCollection|}} + \noiseLevel|\theLanguageCollection|$ for regular languages,
$\theAlphabetSize^{3s|\theLanguageCollection|((s|\theLanguageCollection|)^2+1)^2} + \noiseLevel|\theLanguageCollection|$ for LTT,
and $\theAlphabetSize^{|\theLanguageCollection|(s-1)+1} + \noiseLevel|\theLanguageCollection|$ for PO languages. For context-free languages, the bound remains not computable.
We give upper and lower bounds for all the language classes we study in \cref{sec:noisy-gen-bounds-app}. 

The lower bounds match some, but not all, terms of the upper bounds. For regular and LTT languages, we extend the noiseless lower-bound constructions of
\citet{arenas2026languagegenerationcomplexitybarriers}. For regular languages, we give collections with
$|\theLanguageCollection|=2^\ell$, whose languages are recognized
by automata with at most $c\cdot\ell\cdot n$ states, that require more than
$2^{n^{|\theLanguageCollection|}}+\noiseLevel|\theLanguageCollection|$
examples. The upper and lower bounds exhibit the same nested-exponential structure and matching additive noise terms, but do not match in all parameters. For LTT languages, we give a family of two-language collections with profiles of size $\Theta(n)$ over an alphabet of $n$ symbols
that require more than $2^n+2\noiseLevel$ examples. Compared with the upper bound of $n^{\Theta(n^5)}+2\noiseLevel$, this leaves
a substantial gap in the exponent. For PO languages, we reuse the construction from
\cref{sec:po} and obtain a lower bound of
$\binom{2s}{s}-1+2\noiseLevel$, matching the additive noise term of $2\noiseLevel$ in the upper bound.

\section{Sample complexity of uniform prompted generation}
\label{sec:prompted}

We study how conditioning on a prefix affects the sample complexity of uniform generation. Rather than treating prompted generation as multiclass classification, where each example is labeled with a prompt \citep{raman-generation-plmr}, we follow \citet{kleinberg-2024-language-generation} and treat a prompt as a prefix that the output continues. Thus, the prompt $\thePrompt \in \thePromptDomain$ selects a language via the projection $\theLanguage^\thePrompt = \{\theString \in \theAlphabet^\ast \colon \thePrompt \cdot \theString \in \theLanguage\}$, i.e., the set of valid continuations of $\thePrompt$ in $\theLanguage$. The machinery of \citet{raman-generation-plmr} for uniform prompted generation therefore does not transfer directly; we build it again for our setting below.\looseness=-1

\begin{definition}
    For any collection of languages $\theLanguageCollection$, a sequence of examples $\theStringSet_\theNumExamples=\theString_1,\theString_2, \ldots, \theString_\theNumExamples$, and a prompt $\thePrompt\in\thePromptDomain$, let $\theLanguageCollection(\theStringSet_\theNumExamples,\thePrompt) = \{\theLanguage\in\theLanguageCollection\colon \theStringSet_\theNumExamples \subseteq \theLanguage \text{ and } \theLanguage^\thePrompt \neq \emptyset\}$
    be the set of languages consistent with the enumeration.
    We define the \defn{prompted closure} $\left<\cdot,\cdot\right>_\theLanguageCollection$ as
    \begin{equation}
        \left<\theStringSet_\theNumExamples,\thePrompt\right>_\theLanguageCollection=
        \begin{cases}\bigcap_{\theLanguage\in\theLanguageCollection (\theStringSet_\theNumExamples, \thePrompt)} \theLanguage^\thePrompt, & \text{if } \theLanguageCollection (\theStringSet_\theNumExamples, \thePrompt) \neq \emptyset \\
        \bot, &\text{if } \theLanguageCollection (\theStringSet_\theNumExamples, \thePrompt) = \emptyset.
        \end{cases}
    \end{equation}
\end{definition}

Without a guarantee that a prompt admits arbitrarily many continuations, a
learner can be trapped with no new string to output. To forbid this, we adopt the robustness condition of
\citet{kleinberg-2024-language-generation}.\looseness=-1

\begin{assumption}\label{assumption:infinite-prompted-set}
    For every language $\theLanguage\in\theLanguageCollection$ and every prompt $\thePrompt\in\thePromptDomain$, the language $\theLanguage^\thePrompt$ is infinite.
\end{assumption}

The language $\theLanguage^\thePrompt$ is the standard left quotient of $\theLanguage$ by $\thePrompt$, and whether it is infinite is decidable for all language classes we study. For instance, for PO languages, the left quotient is recognized by the automaton obtained by shifting the initial states to those reachable by reading $\thePrompt$, preserving the partial order structure and generally leaving the number of states unchanged.

Formal definitions, a characterization of prompted uniformly generatable collections, and all proofs are deferred to \cref{sec:prompted-gen-bounds-app}. Here, we use only the notion of $\distinctNumExamples$-prompted uniform generatability: a collection is $\distinctNumExamples$-prompted uniformly generatable if, for every prompt $\thePrompt$, uniform generation succeeds after observing $\distinctNumExamples$ distinct examples continuing $\thePrompt$.\looseness=-1

\begin{restatable}{corollary}{PromptedGenFiniteCollProp}
    Let $\thePromptDomain$ be finite.
    Any finite collection $\theLanguageCollection$ of infinite languages satisfying \cref{assumption:infinite-prompted-set} is prompted uniformly generatable.
\end{restatable}

For general collections, the finiteness of the prompt domain $\thePromptDomain$ is essential. When $\thePromptDomain$ is infinite, even a collection of two languages may fail to be prompted uniformly generatable (see \cref{sec:prompted-gen-bounds-app}). For regular, LTT and PO languages, however, the left quotient by any prompt has a representation of bounded size, so the bounds do not depend on the prompt and are valid even when $\thePromptDomain$ is infinite.

\begin{restatable}{proposition}{PropPromptedUnifGen}
\label{prop:prompted-closure-dim}
    A finite collection $\theLanguageCollection$ of languages satisfying \cref{assumption:infinite-prompted-set} is \emph{not} $\distinctNumExamples$-prompted uniformly generatable if and only if there exist a prompt $\thePrompt$ and a non-empty set $\theLanguageCollSubset \subseteq \theLanguageCollection$ such that $\bigcap_{\theLanguage \in \theLanguageCollSubset} \theLanguage^\thePrompt$ is finite and has size at least $\distinctNumExamples$.
\end{restatable}

\begin{restatable}{theorem}{ThmPromptedGeneric}
\label{thm:prompt-generic}
Let $\theLanguageCollection$ be a finite collection of infinite languages satisfying \cref{assumption:infinite-prompted-set}, and let $\distinctNumExamples = \sup_{\thePrompt \in \thePromptDomain} B(\{\theLanguage^\thePrompt \colon \theLanguage \in \theLanguageCollection\})$. If $\distinctNumExamples < \infty$, then $\theLanguageCollection$ is $\distinctNumExamples$-prompted uniformly generatable.
\end{restatable}

\begin{proof}[Proof sketch]
    By \cref{prop:prompted-closure-dim}, the sample complexity is governed by the largest finite intersection $\bigcap_{\theLanguage \in \theLanguageCollSubset} \theLanguage^\thePrompt$ over all prompts $\thePrompt \in \thePromptDomain$ and non-empty $\theLanguageCollSubset \subseteq \theLanguageCollection$. For a fixed prompt $\thePrompt$, this is exactly the quantity the recipe of \cref{sec:recipe} bounds, applied to the collection of left quotients $\{\theLanguage^\thePrompt \colon \theLanguage \in \theLanguageCollection\}$ rather than to $\theLanguageCollection$ itself. \cref{assumption:infinite-prompted-set} guarantees the quotients are infinite, so the recipe is applicable. Taking the supremum over prompts gives $\distinctNumExamples$. The full proof is in \cref{sec:prompted-gen-bounds-app}.\looseness=-1
\end{proof}

The bound therefore depends only on how the left quotient affects the language representation: for regular and PO languages it is unchanged, since the quotient is recognized by an automaton with at most as many states as the original, for LTT languages, the quotient is a finite union of profiles of less than twice the original size, so the profile size at most doubles, independently of $\thePrompt$, and for context-free languages, uncomputability is preserved. All constructions and bounds are given in
\cref{sec:prompted-gen-bounds-app}.\looseness=-1

\section{Sample complexity of uniform generation with finite feedback}
\label{sec:feedback}

\citet{Bai_2026} prove that generation in the limit with finitely many feedback queries is equivalent in power to generation in the limit without feedback. For uniform generation, however, feedback can make a difference: \citet{charikar2024exploringfacetslanguagegeneration} give a collection that is uniformly generatable with a single query but not without feedback. We show that, for finite collections, a generator with a query budget of $\theNumQueries$ can sometimes achieve uniform generation after seeing fewer examples. We follow \citet{Bai_2026} in allowing a total budget of $\theNumQueries$ queries, assuming $|\theLanguageCollection| - 1 > \theNumQueries$, since otherwise queries alone suffice to eliminate all but one language and generate without any examples. This differs from \citet{charikar2024exploringfacetslanguagegeneration}, who allow one query per timestep. We define some analogous notions.\looseness=-1

\begin{definition}
For any collection of languages $\theLanguageCollection$, an enumeration $\theStringSet_\theNumExamples = \theString_1, \theString_2, \ldots, \theString_\theNumExamples$, and a sequence of queries $Q_k = \theQueryString_1, \ldots, \theQueryString_k$ such that $k \leq \theNumQueries$, answered according to the target language $\theTrueLanguage \in \theLanguageCollection$, let
\begin{equation}
    \theLanguageCollection(\theStringSet_\theNumExamples, Q_k) = \{\theLanguage \in \theLanguageCollection \colon \theStringSet_\theNumExamples \subseteq \theLanguage \text{ and } \mathbbm{1}\{\theQueryString_i \in \theLanguage\} = \mathbbm{1}\{\theQueryString_i \in \theTrueLanguage\}, \forall i \in [1, k]\}
\end{equation}
be the set of languages consistent with the enumeration and the queries' answers.
We define the \defn{feedback-augmented closure} $\left<\cdot, \cdot\right>_\theLanguageCollection$ as
\begin{equation}
    \left<\theStringSet_\theNumExamples, Q_k\right>_\theLanguageCollection =
    \begin{cases}
        \bigcap_{\theLanguage \in \theLanguageCollection(\theStringSet_\theNumExamples, Q_k)} \theLanguage, & \text{if } \theLanguageCollection(\theStringSet_\theNumExamples, Q_k) \neq \emptyset, \\
        \bot, & \text{if } \theLanguageCollection(\theStringSet_\theNumExamples, Q_k) = \emptyset.
    \end{cases}
\end{equation}
\end{definition}

We adapt the definition of uniform generatability with feedback and the characterization of uniformly generatable collections from \citet{charikar2024exploringfacetslanguagegeneration} to our fixed-budget setting. We say a collection is $\distinctNumExamples$-generatable with $\theNumQueries$ feedback queries if uniform generation succeeds after observing $\distinctNumExamples$ distinct examples, using at most $\theNumQueries$ queries in total. The characterization below reduces the sample complexity in this setting to the size of a closure, which is what allows us to derive the bounds. Formal definitions and proofs are given in \cref{sec:feedback-gen-bounds-app}.

\begin{restatable}{proposition}{PropUnifGenFeedback}\label{prop:unif-gen-feedback}
A finite collection $\theLanguageCollection$ of infinite languages is not $\distinctNumExamples$-generatable with $\theNumQueries$ feedback queries if and only if for every generator $\learner$ that asks at most $\theNumQueries$ feedback queries, there exist a target $\theTrueLanguage \in \theLanguageCollection$, an enumeration $\theCurrentStringSet = \theString_1, \ldots, \theString_\theTimestep$ of $\theTrueLanguage$ with $|\{\theString_1, \ldots, \theString_\theTimestep\}| \geq \distinctNumExamples$, and the answers according to $\theTrueLanguage$ to the queries $Q_k$ asked by $\learner$ up to and including timestep $\theTimestep$, such that $\langle \theCurrentStringSet, Q_k \rangle_\theLanguageCollection$ is finite, has size at least $\distinctNumExamples$, and contains only strings of $\theCurrentStringSet$.
\end{restatable}

\paragraph{The role of feedback queries.} 
By \cref{prop:unif-gen-feedback}, the sample complexity is determined by whether the feedback-augmented closure can be exhausted by the observed examples, and therefore by the size of the largest finite feedback-augmented closure. Feedback queries help by eliminating languages from the consistent set, potentially making the intersection infinite. If the query budget suffices to eliminate the right languages regardless of query answers, the generator needs no examples at all. We show that by choosing informative queries that eliminate languages from the consistent set, the closure size, and hence the sample complexity, can sometimes be reduced substantially. A simple strategy eliminates one language per query, reducing the effective collection size by $\theNumQueries$. One might hope to do better by halving the consistent set with each query, but such queries may not exist, as we show below.\looseness=-1

\paragraph{A simple strategy.} 
Given a query budget $\theNumQueries$, the generator picks any two languages $\theLanguage_1,\theLanguage_2$ in the consistent set and queries a string $\theQueryString\in\theLanguage_1\triangle\theLanguage_2=(\theLanguage_1\setminus\theLanguage_2) \cup (\theLanguage_2\setminus\theLanguage_1)$.
Since all languages in $\theLanguageCollection$ are distinct, such a string always exists, and regardless of the answer, at least one language is eliminated. Repeating this $\theNumQueries$ times leaves at most $|\theLanguageCollection|-\theNumQueries$ languages in the consistent set. We give the following generic theorem, and we instantiate it for regular, LTT and PO languages, and prove uncomputability for context-free languages in \cref{sec:feedback-gen-bounds-app}.

\begin{restatable}{theorem}{ThmFeedbackSimple}
\label{thm:feedback-generic}
Let $\theLanguageCollection$ be a finite collection of infinite languages and let $\theNumQueries$ be the feedback budget. Then $\theLanguageCollection$ is $\max\{B(\theLanguageCollection') \colon \theLanguageCollection' \subseteq \theLanguageCollection, |\theLanguageCollection'| = |\theLanguageCollection| - \theNumQueries\}$-generatable with $\theNumQueries$ feedback queries.
\end{restatable}

\begin{proof}[Proof sketch]
Feedback helps by shrinking the
consistent set. For each of the $\theNumQueries$ queries, the generator picks two
languages $\theLanguage_1, \theLanguage_2$ from the current consistent set and
queries a string in $\theLanguage_1 \triangle \theLanguage_2$, which is non-empty
since the languages are distinct. Whatever the answer is, at least one of them is
eliminated. After $\theNumQueries$ queries, at most $|\theLanguageCollection| -
\theNumQueries$ languages remain consistent, and the recipe of
\cref{sec:recipe} applied to that smaller collection gives the bound. The full
proof is in \cref{sec:feedback-gen-bounds-app}.\looseness=-1
\end{proof}

\paragraph{A better strategy.} 
Instead of separating two languages at a time, one could try to halve the consistent set $C$ with every query. Each query $\theQueryString$ should then split the consistent set into the languages that contain $\theQueryString$ and those that do not, and whichever way the oracle answers, one of the two parts is eliminated. If some query splits the consistent set into two roughly equal parts, for instance a string $\theQueryString \in (\bigcap_{\theLanguage \in C_1} \theLanguage) \setminus (\bigcup_{\theLanguage \in C_2} \theLanguage)$ for halves $C_1, C_2$, a single query reduces the consistent set to about half, and $\theNumQueries$ queries reduce it to about $|C|/2^\theNumQueries$. However, such a query need not exist, in which case this strategy fails. We give an example below for LTT languages.

\begin{restatable}{theorem}{LTTFeedbackGenThmLowerBound}
There exists a finite collection of $n$ infinite LTT languages $\theLanguageCollection$ over an alphabet of size $\theAlphabetSize=n\geq 4$, such that $\theLanguageCollection$ is \emph{not} $2^n$-generatable with $\theNumQueries$ feedback queries, for any $\theNumQueries$ such that $\theNumQueries<|\theLanguageCollection|-1$.
\end{restatable}

\begin{proofsketch}
    Let $\theLanguageCollection$ consist of $n$ LTT languages over $\theAlphabet=\{1,\ldots, n\}$, for $n \ge 4$, and $\theLanguage_i=(\varepsilon,\varepsilon,\{(j, =1)\colon j\neq i\})$, for all $i$. $\theLanguage_i$ constrains every symbol except $i$ to appear exactly once, and is infinite since $i$ can appear any number of times. For any subset $\theLanguageCollSubset\subseteq \theLanguageCollection$ with $|\theLanguageCollSubset|\geq 2$, every symbol is constrained in at least one language, forcing all $n$ symbols to appear once. The intersection is therefore always the same: all permutations of $\{1,\ldots, n\}$, of size $n!\ge 2^n$ for $n\ge 4$. No query eliminates more than one language: a string whose symbols all occur exactly once belongs to every $\theLanguage_i$, a string where exactly one symbol $i$ occurs a different number of times belongs to $\theLanguage_i$ only, and any other string belongs to none. Since $\theNumQueries<|\theLanguageCollection|-1$, at least two languages remain consistent and the closure is the set of permutations.
    Therefore, $\theLanguageCollection$ is not $2^n$-generatable with $\theNumQueries$ feedback queries, for any $\theNumQueries$ such that $\theNumQueries<|\theLanguageCollection|-1$.
\end{proofsketch}

Selecting informative queries requires computing differences, and hence
complement, which for NFAs incurs an exponential blowup via determinization
\citep{hopcroft-2006-introduction}. However, this overhead
affects the query selection only.
Moreover, the symmetric difference of PO languages may not be PO, since PO languages are not closed under complement.
Once languages are
eliminated, the remaining collection is bounded using its original
representation, so the sample complexity bounds are unaffected. \looseness=-1

\section{Discussion and conclusion}

We studied the sample complexity of uniform language generation for context-free,
regular, LTT, and PO languages, in
extensions of the language generation in the limit framework that model noisy input, prompting, and feedback. We found that PO languages, whose sample complexity had not
been studied before, require an exponential number of
samples. A recurring theme is that finite noise adds only a linear term, prompting leaves the bounds unchanged up to a constant factor in the exponent, and finite feedback can reduce them exponentially, while the bounds for context-free languages remain uncomputable regardless of the setting.

Some of our assumptions are strong, and a language model trained on a real corpus may not satisfy them. However, these assumptions are reasonable in scenarios such as constrained generation. Here, outputs must be valid
against a schema or grammar, membership is indeed binary, there are only a few, fixed
candidate languages, and the oracle is a validator rather than
a human, so feedback is truthful and cheap. Our bounds are
exponential in $|\theLanguageCollection|$,  so they are meaningful only when the
collection is small. The exponent therefore matters: the smaller one we give for PO languages allows a considerably larger collection before the bound becomes vacuous.
The uncomputability result is the most consequential: for languages at least as expressive as context-free languages, a uniform bound exists but no algorithm can compute it for any such collection, so one can never know in
advance how many examples suffice.\looseness=-1

Finally, our results should be read together with recent work on mistake-bounded
generation, which answers a different but complementary question. Our results bound the time after which the learner is guaranteed to be correct. In contrast, \citet{kleinberg2026mistakeboundedlanguagegeneration} bound how many mistakes the learner makes in total, showing that for any
finite collection there is a learner that makes at most $\min\{\lfloor \log_2 |\theLanguageCollection| \rfloor,
\closureDim(\theLanguageCollection)\}$ mistakes. Taken together, these results say that the learner may make mistakes rarely, but need many examples before it never makes mistakes again. Moreover, these results do not consider the learner's computation. \citet{jimenez2026polynomialtimemistakeboundedlanguagegeneration} show that, under a standard cryptographic assumption, some collections do not admit efficient learners with few mistakes.

\bibliography{references}

\begin{thebibliography}{19}
\providecommand{\natexlab}[1]{#1}
\providecommand{\url}[1]{\texttt{#1}}
\expandafter\ifx\csname urlstyle\endcsname\relax
  \providecommand{\doi}[1]{doi: #1}\else
  \providecommand{\doi}{doi: \begingroup \urlstyle{rm}\Url}\fi

\bibitem[Angluin(1980)]{ANGLUIN1980}
Dana Angluin.
\newblock Inductive inference of formal languages from positive data.
\newblock \emph{Information and Control}, 45\penalty0 (2):\penalty0 117--135, 1980.
\newblock URL \url{https://www.sciencedirect.com/science/article/pii/S0019995880902855}.

\bibitem[Arenas et~al.(2026)Arenas, Barcelo, Cofr{\'e}, and Kozachinskiy]{arenas2026languagegenerationcomplexitybarriers}
Marcelo Arenas, Pablo Barcelo, Luis Cofr{\'e}, and Alexander Kozachinskiy.
\newblock Language generation in the limit: Complexity barriers and implications for learning.
\newblock In \emph{Forty-third International Conference on Machine Learning}, 2026.
\newblock URL \url{https://openreview.net/forum?id=3I6v86xf2l}.

\bibitem[Bai et~al.(2026)Bai, Panigrahi, and Zhang]{Bai_2026}
Yannan Bai, Debmalya Panigrahi, and Ian Zhang.
\newblock Language generation in the limit: Noise, loss, and feedback.
\newblock In \emph{Proceedings of the 2026 Annual {ACM-SIAM} Symposium on Discrete Algorithms}, pp.\  794--816, 2026.
\newblock URL \url{https://doi.org/10.1137/1.9781611978971.31}.

\bibitem[Brzozowski \& Fich(1980)Brzozowski and Fich]{BRZOZOWSKI198032}
J.A. Brzozowski and Faith~E. Fich.
\newblock Languages of r-trivial monoids.
\newblock \emph{Journal of Computer and System Sciences}, 20\penalty0 (1):\penalty0 32--49, 1980.
\newblock ISSN 0022-0000.
\newblock \doi{https://doi.org/10.1016/0022-0000(80)90003-3}.
\newblock URL \url{https://www.sciencedirect.com/science/article/pii/0022000080900033}.

\bibitem[Charikar \& Pabbaraju(2024)Charikar and Pabbaraju]{charikar2024exploringfacetslanguagegeneration}
Moses Charikar and Chirag Pabbaraju.
\newblock Exploring facets of language generation in the limit, 2024.
\newblock URL \url{https://arxiv.org/abs/2411.15364}.

\bibitem[Gold(1967)]{gold1967language}
E~Mark Gold.
\newblock Language identification in the limit.
\newblock \emph{Information and Control}, 10\penalty0 (5):\penalty0 447--474, 1967.
\newblock URL \url{https://www.sciencedirect.com/science/article/pii/S0019995867911655}.

\bibitem[Hella \& Vilander(2022)Hella and Vilander]{hella2022}
Lauri Hella and Miikka Vilander.
\newblock Defining long words succinctly in {FO} and {MSO}.
\newblock In \emph{Revolutions and Revelations in Computability: 18th Conference on Computability in Europe}, pp.\  125–138, 2022.
\newblock URL \url{https://doi.org/10.1007/978-3-031-08740-0_11}.

\bibitem[Hopcroft et~al.(2006)Hopcroft, Motwani, and Ullman]{hopcroft-2006-introduction}
John~E. Hopcroft, Rajeev Motwani, and Jeffrey~D. Ullman.
\newblock \emph{Introduction to Automata Theory, Languages, and Computation}.
\newblock 3rd edition, 2006.
\newblock URL \url{https://dl.acm.org/doi/book/10.5555/1196416}.

\bibitem[Jimenez et~al.(2026)Jimenez, Kozachinskiy, and Opazo]{jimenez2026polynomialtimemistakeboundedlanguagegeneration}
Héctor Jimenez, Alexander Kozachinskiy, and Vicente Opazo.
\newblock Polynomial-time mistake-bounded language generation, 2026.
\newblock URL \url{https://arxiv.org/abs/2606.16077}.

\bibitem[Kalavasis et~al.(2025)Kalavasis, Mehrotra, and Velegkas]{kalavasis2025limits}
Alkis Kalavasis, Anay Mehrotra, and Grigoris Velegkas.
\newblock On the limits of language generation: Trade-offs between hallucination and mode-collapse.
\newblock In \emph{Proceedings of the 57th Annual ACM Symposium on Theory of Computing}, STOC '25, pp.\  1732–1743, New York, NY, USA, 2025. Association for Computing Machinery.
\newblock ISBN 9798400715105.
\newblock \doi{10.1145/3717823.3718108}.
\newblock URL \url{https://doi.org/10.1145/3717823.3718108}.

\bibitem[Kleinberg \& Mullainathan(2024)Kleinberg and Mullainathan]{kleinberg-2024-language-generation}
Jon Kleinberg and Sendhil Mullainathan.
\newblock Language generation in the limit.
\newblock In \emph{Advances in Neural Information Processing Systems}, volume~37, pp.\  66058--66079, 2024.
\newblock URL \url{https://proceedings.neurips.cc/paper_files/paper/2024/file/7988e9b3876ad689e921ce05d711442f-Paper-Conference.pdf}.

\bibitem[Kleinberg et~al.(2026)Kleinberg, Peale, and Reingold]{kleinberg2026mistakeboundedlanguagegeneration}
Jon Kleinberg, Charlotte Peale, and Omer Reingold.
\newblock Mistake-bounded language generation, 2026.
\newblock URL \url{https://arxiv.org/abs/2605.10809}.

\bibitem[Li \& Zhang(2026)Li and Zhang]{li2026characterizing}
Aaron Li and Ian Zhang.
\newblock Characterizing the effect of noise in language generation in the limit.
\newblock In \emph{Forty-third International Conference on Machine Learning}, 2026.
\newblock URL \url{https://openreview.net/forum?id=bo3bKNtIsI}.

\bibitem[Li et~al.(2025)Li, Raman, and Tewari]{raman-generation-plmr}
Jiaxun Li, Vinod Raman, and Ambuj Tewari.
\newblock Generation through the lens of learning theory.
\newblock In \emph{Proceedings of Thirty Eighth Conference on Learning Theory}, volume 291 of \emph{Proceedings of Machine Learning Research}, pp.\  4740--4776, 2025.
\newblock URL \url{https://proceedings.mlr.press/v291/raman25a.html}.

\bibitem[Li et~al.(2026)Li, Han, Shi, Jiang, and Gao]{li2026computationalhardnessmistakeboundedlanguage}
Xiaoyu Li, Andi Han, Dai Shi, Jiaojiao Jiang, and Junbin Gao.
\newblock On computational hardness of mistake-bounded language generation: A random-oracle query separation, 2026.
\newblock URL \url{https://arxiv.org/abs/2608.05029}.

\bibitem[Masopust \& Krötzsch(2021)Masopust and Krötzsch]{Masopust_2021}
Tomáš Masopust and Markus Krötzsch.
\newblock Partially ordered automata and piecewise testability.
\newblock \emph{Logical Methods in Computer Science}, Volume 17, Issue 2, May 2021.
\newblock ISSN 1860-5974.
\newblock \doi{10.23638/lmcs-17(2:14)2021}.
\newblock URL \url{http://dx.doi.org/10.23638/LMCS-17(2:14)2021}.

\bibitem[Mehrotra et~al.(2025)Mehrotra, Velegkas, Yu, and Zhou]{mehrotra2025languagegenerationinfinitecontamination}
Anay Mehrotra, Grigoris Velegkas, Xifan Yu, and Felix Zhou.
\newblock Language generation with infinite contamination, 2025.
\newblock URL \url{https://arxiv.org/abs/2511.07417}.

\bibitem[Raman \& Raman(2025)Raman and Raman]{raman2025generation}
Ananth Raman and Vinod Raman.
\newblock Generation from noisy examples.
\newblock In \emph{Forty-second International Conference on Machine Learning}, 2025.
\newblock URL \url{https://openreview.net/forum?id=ouaiJlE1hJ}.

\bibitem[Schwentick et~al.(2001)Schwentick, Th{\'e}rien, and Vollmer]{Schwentick-2001-partially}
Thomas Schwentick, Denis Th{\'e}rien, and Heribert Vollmer.
\newblock Partially-ordered two-way automata: A new characterization of da.
\newblock In \emph{Revised Papers from the 5th International Conference on Developments in Language Theory}, DLT '01, pp.\  239–250, Berlin, Heidelberg, 2001. Springer-Verlag.
\newblock ISBN 3540434534.

\end{thebibliography}
\bibliographystyle{iclr2027_conference}

\newpage
\appendix

\section{Related work}
\label{sec:related-work}

\paragraph{Generation vs. prediction.}
\Citet{raman-generation-plmr} compare uniform language generation in the limit with classic prediction frameworks such as PAC and online learning. In these frameworks, learnability is characterized by the finiteness of a combinatorial dimension: the VC dimension for PAC learnability, the Littlestone dimension for online learnability. Uniform generation is similar, and the dimension characterizing it is the closure dimension we use in this paper. The authors show that generation and prediction are incomparable: there are classes that are uniformly generatable but neither PAC nor online learnable, and classes that are PAC or online learnable but not generatable. Generation is therefore a different problem, and the
sample complexity cannot be read off from these standard dimensions.\looseness=-1

\paragraph{Mistake-bounded generation.} 
A recent line of work studies mistake-bounded language generation \citep{kleinberg2026mistakeboundedlanguagegeneration, jimenez2026polynomialtimemistakeboundedlanguagegeneration, li2026computationalhardnessmistakeboundedlanguage}, which counts how many mistakes the learner makes in total rather than when it makes the last mistake. For finite collections, \cite{kleinberg2026mistakeboundedlanguagegeneration} give a worst-case tight bound on the number of mistakes, and show that under adversarial noise the number of mistakes degrades at most linearly in the number of noisy examples. \cite{jimenez2026polynomialtimemistakeboundedlanguagegeneration} introduce a polynomial-time version of the setting and show that some collections admit polynomial-time generators with polynomially many mistakes, while others do not under the strong RSA assumption. \cite{li2026computationalhardnessmistakeboundedlanguage} show a larger gap in a different model. They give a collection of infinite languages on which a computationally unbounded generator makes no mistakes, while every polynomial-time generator makes exponentially many expected mistakes in the worst case.

\paragraph{Statistical formulations.}
A different line of work replaces the enumeration with a distribution.
\citet{kalavasis2025limits} study generation from random samples drawn from a
distribution over an unknown target, and ask for consistency and
breadth, i.e., as the size of the
revealed set increases, the set of strings the learner outputs converges to
the set of all unseen strings in the target language.
They show that this requirement is much stronger than that of \citet{kleinberg-2024-language-generation}:
it is unachievable for most collections and models,
including next-token predictors.\looseness=-1

\section{Further preliminaries}
\subsection{Context-free languages}
\label{sec:cf}

\begin{definition}
    A \defn{context-free grammar} (CFG) is a tuple $\theGrammar=\theGrammarTuple$, where $\theVariables$ is a finite set of \defn{nonterminal symbols}, $\theAlphabet$ is a finite set of \defn{terminal symbols}, $\theRules \subseteq \theVariables \times (\theVariables \cup \theAlphabet)^\ast$ is a finite set of \defn{production rules}, and $\theStartVariable \in \theVariables$ is a distinguished \defn{start symbol}. 
\end{definition}

Using production rules, we can \defn{derive} new symbols by replacing the variable of the left-hand side with the sequence of symbols on the right-hand side, i.e.,  $\alpha_1 \theVariable\alpha_2 \Rightarrow_\theGrammar \alpha_1 \theRHS\alpha_2$, where $\theVariable\rightarrow\theRHS\in\theRules$ and $\alpha_1,\alpha_2\in(\theVariables \cup \theAlphabet)^\ast$. A CFG $\theGrammar$ \defn{derives} a string $\theString\in\theAlphabet^\ast$ from the start symbol $\theStartVariable$ by repeatedly deriving symbols, until only terminal symbols are left, i.e., $\theStartVariable\xRightarrow[\theGrammar]{\ast}\theString$. The language generated by $\theGrammar$ is $L(\theGrammar)=\{\theString\in\theAlphabet^\ast\mid\theStartVariable\xRightarrow[\theGrammar]{\ast}\theString\}$. A language is \defn{context-free} if it is generated by some CFG.

\begin{example}
    Consider the alphabet $\{a,b\}$. The language $\theLanguage=\{a^n b^n\mid n \ge 0\}$, consisting of all strings with equal number of $a$s and $b$s, where all $a$s precede the $b$s, is context-free, and it is generated by the grammar $\theGrammar=(\{S\}, \theAlphabet,\theRules,S)$, with production rules $\theRules=\{S\rightarrow\varepsilon,S\rightarrow a S b\}$. 
\end{example}

\subsection{Turing machines and the halting problem}\label{sec:turing-machines}

\begin{definition}
    A \defn{Turing machine} is a tuple $\mathcal{M}=(\theStates,\Gamma, \theTransitions,\theInitState,\theState_F)$, where $\theStates$ is a finite set of states, $\Gamma$ is a finite alphabet of tape symbols, including a blank symbol $\bot$, $\theTransitions\colon \theStates\times\Gamma\rightarrow \theStates\times \Gamma\times \{L,R\}$ is a transition function, $\theInitState$ is the initial state, and $\theState_F$ is the final state.
\end{definition}
A \defn{configuration} of $\mathcal{M}$ is a triple $(\theState,\gamma,i)$, indicating the current state $\theState$, the current tape contents $\gamma\in\Gamma^\ast$, and the position $i$ of the head on the tape. A \defn{computation} of $\mathcal{M}$ is a sequence of configurations $C_0,C_1,\ldots$ such that $C_0=(\theInitState, \bot, 0)$ is the initial configuration and $C_{n+1}$ is obtained from $C_n$ by applying $\theTransitions$. We say that $\mathcal{M}$ \defn{halts} if it reaches a configuration such that the current state is the final state. 

The \defn{halting problem} asks whether a Turing machine $\mathcal{M}$ ever halts. It is a classic result that the halting problem is undecidable \citep{hopcroft-2006-introduction}. More specifically, there is no algorithm that determines whether a Turing machine halts. We use this fact throughout this paper to establish the uncomputability of several sample complexity bounds. Specifically, we show that the existence of a computable bound would allow us to simulate a Turing machine for a finite number of steps and correctly decide whether it halts, contradicting undecidability.

\section{Sample complexity of partially-ordered languages}\label{sec:app-piecewise-testable}

\TheoremLTBounds*
\begin{proof}
    It suffices to prove that for any non-empty subset $\theLanguageCollSubset\subseteq\theLanguageCollection$, if the intersection $\bigcap_{\theLanguage\in\theLanguageCollSubset}\theLanguage$ is finite, then the size of this intersection is at most our desired bound. Consider the product automaton $\theAutomaton$ encoding the intersection of each $\theAutomaton_\theLanguage$, for $\theLanguage\in\theLanguageCollSubset$. We assumed that each $\theAutomaton_\theLanguage$ is recognized by a PONFA with at most $s$ states. The length of the longest accepting path of each PONFA is at most $s-1$. By definition, each $\theAutomaton_\theLanguage$ is partially-ordered, therefore $\theAutomaton$ is also partially-ordered. Its states are of the form $(\theState_1, \ldots, \theState_{|\theLanguageCollSubset|})$ and it holds that $(\theState_1,\ldots,\theState_n)\leq (\theState'_1,\ldots,\theState'_n)$ if $\theState_i \leq \theState'_i$ for all $i$. Since every accepting path in $\theAutomaton$ is simple, the length of any accepted string is bounded by the length of the longest simple path in $\theAutomaton$. We now bound this length using the partial order structure of $\theAutomaton$. At each step of a simple path, the tuple of states must strictly increase in at least one component, otherwise the same tuple would be visited twice, contradicting the definition of simple path. Since each $\theAutomaton_\theLanguage$ has at most s states, it can strictly increase at most $s-1$ times. Since there are $|\theLanguageCollSubset|\leq |\theLanguageCollection|$ PONFAs,  the total number of strict increases is at most $|\theLanguageCollection|(s-1)$. Therefore the longest simple path in $\theAutomaton$ has length at most $|\theLanguageCollection|(s-1)$, meaning every string has length at most $|\theLanguageCollection|(s-1)$. Since $\theAutomaton$ accepts only strings of length at most $|\theLanguageCollection|(s-1)$, the intersection contains less than $\theAlphabetSize^{|\theLanguageCollection|(s-1)+1}$ strings. 
\end{proof}

\begin{restatable}{theorem}{POLangLowerBound}
    For every integer $s\geq 2$, there exists a collection
    $\theLanguageCollection$ of two infinite PO languages over a
    binary alphabet, each recognized by a PONFA with $s$ states, such that $\theLanguageCollection$ is not $\left(\binom{2s}{s}-1\right)$-generatable.
\end{restatable}

\begin{proof}
    Let $\theAlphabet=\{a,b\}$ and define
    $\theLanguage_1=\{w\in\theAlphabet^\ast : \#_a(w)\leq s-1\}$
    and
    $\theLanguage_2=\{w\in\theAlphabet^\ast : \#_b(w)\leq s-1\}$,
    where $\#_a(w)$ and $\#_b(w)$ denote the numbers of occurrences
    of $a$ and $b$ in $w$, respectively. Both languages are infinite.

    A PONFA recognizing $\theLanguage_1$ has states
    $q_0,\ldots,q_{s-1}$, with $q_0$ initial and all states
    accepting. Each state has a self-loop on $b$, and there is
    a transition on $a$ from $q_i$ to $q_{i+1}$ for
    $0\leq i<s-1$, with no transition on $a$ from $q_{s-1}$.
    Exchanging $a$ and $b$ gives a PONFA recognizing
    $\theLanguage_2$. These automata generalize those in
    \cref{fig:ponfa-example} and are partially-ordered.

    The intersection $\theLanguage_1\cap\theLanguage_2$ consists
    exactly of the strings with at most $s-1$ occurrences of
    each symbol, so it is finite. Since there are
    $\binom{i+j}{i}$ strings containing exactly $i$ occurrences
    of $a$ and $j$ occurrences of $b$, its size is
    $|\theLanguage_1\cap\theLanguage_2|
    =\sum_{i=0}^{s-1}\sum_{j=0}^{s-1}\binom{i+j}{i}
    =\sum_{i=0}^{s-1}\binom{i+s}{i+1}
    =\binom{2s}{s}-1$,
    where the last two equalities follow from the hockey-stick
    identity. By \cref{prop:arenas-prop-two-three}, the collection
    $\theLanguageCollection=\{\theLanguage_1,\theLanguage_2\}$
    is not $\left(\binom{2s}{s}-1\right)$-generatable.
    Since $\binom{2s}{s}\sim 4^s/\sqrt{\pi s}$ by Stirling's approximation, while \cref{thm:lt-bound} gives the upper bound $2^{2(s-1)+1}=4^s/2$ for these collections, the two differ by a factor of $\Theta(\sqrt{s})$.
\end{proof}

\section{Uniform finite-noise generation}\label{sec:noisy-gen-bounds-app}

\begin{definition}\label{def:unif-lang-gen}
    Let $\noiseLevel \in \mathbb{N}$ be a finite noise level. A language collection $\theLanguageCollection$ is \defn{uniformly $\distinctNumExamples$-generatable at noise level $\noiseLevel$} if there exists a generator $\learner$ such that for every $\theLanguage \in \theLanguageCollection$ and every enumeration $\theString_1, \theString_2, \ldots$ of $\theLanguage$ with $\sum_{t=1}^\infty \mathbbm{1}\{\theString_t \notin \theLanguage\} \leq \noiseLevel$, if there exists $\theTimestepStar \in \mathbb{N}$ such that $|\{\theString_1, \ldots, \theString_\theTimestepStar\}| = \distinctNumExamples$, then $\learner(\theString_1, \ldots, \theString_\theTimestep) \in \theLanguage \setminus \{\theString_1, \ldots, \theString_\theTimestep\}$ for all $\theTimestep \geq \theTimestepStar$.
\end{definition}

We follow \citet{li2026characterizing} and allow the generator to depend on the noise level $\noiseLevel$. We fix the noise level because we are interested in the sample complexity at a given noise level, i.e., the threshold $\distinctNumExamples$ as a function of $\noiseLevel$. With $\noiseLevel$ fixed, $\distinctNumExamples$ is characterized directly by the noisy closure at noise level $\noiseLevel$ (\cref{prop:noisy-unif-gen}), in analogy to the noiseless characterization of \citet{arenas2026languagegenerationcomplexitybarriers}.

\begin{corollary}
    Every finite collection is uniformly $\distinctNumExamples$-generatable at noise level $\noiseLevel$ for some $\distinctNumExamples$, for every $\noiseLevel$.
\end{corollary}
\begin{proof}
    \citet{raman2025generation} show that every finite collection of infinite languages admits a single generator that succeeds at every finite noise level, with a threshold that may depend on the noise level. Fixing the noise level gives the claim.
\end{proof}

\begin{example}
    Take $\theLanguageCollection=\{\theLanguage_1,\theLanguage_2,\theLanguage_3\}$ and $\noiseLevel=1$. Assume $\theStringSet_3=\{x,y,z\}$ such that $x\in\theLanguage_1\cap\theLanguage_2\cap\theLanguage_3$, $y\in\theLanguage_1\cap\theLanguage_2$ and $z\in\theLanguage_2\cap\theLanguage_3$. All three languages are consistent at noise level $\noiseLevel=1$ because $\theLanguage_1$ misses only $z$, $\theLanguage_2$ misses nothing and $\theLanguage_3$ misses only $y$. So the noisy closure is $\theLanguage_1\cap\theLanguage_2\cap\theLanguage_3$.
\end{example}

\NoisyUnifGenerationProposition*

\begin{proof}
    $(\Rightarrow)$ Suppose $\theLanguageCollection$ is not $\distinctNumExamples$-generatable at noise level $\noiseLevel$. Let $\learner^*$ be a generator that outputs an arbitrary unseen element from the noisy closure when one exists, and an arbitrary string otherwise. Since $\learner^*$ must fail, there exist a language $\theLanguage \in \theLanguageCollection$, an enumeration $\theString_1, \theString_2, \ldots$ of $\theLanguage$ with $\sum_t \mathbbm{1}\{\theString_t \notin \theLanguage\} \leq \noiseLevel$, and a timestep $s$ with $|\{\theString_1, \ldots, \theString_s\}| \geq \distinctNumExamples$ such that $\learner^*(\theString_1, \ldots, \theString_s) \notin \theLanguage \setminus \{\theString_1, \ldots, \theString_s\}$. Let $\theStringSet_s = \{\theString_1, \ldots, \theString_s\}$ and $X = \langle \theStringSet_s, \noiseLevel \rangle_\theLanguageCollection$. At most $\noiseLevel$ strings of $\theStringSet_s$ are outside $\theLanguage$, so $\theLanguage \in \theLanguageCollection(\theStringSet_s, \noiseLevel)$ and $X \subseteq \theLanguage$.

\noindent\textit{$X$ is finite:} Otherwise, $X \setminus \theStringSet_s \neq \emptyset$, so $\learner^*$ outputs an element of $X \setminus \theStringSet_s \subseteq \theLanguage \setminus \theStringSet_s$, contradicting that $\learner^*$ fails at time $s$.

\noindent\textit{Reducing to $\distinctNumExamples$ strings:} Let $\theStringSet'$ be an enumeration of $\distinctNumExamples$ distinct strings of $\theStringSet_s$. Every language missing at most $\noiseLevel$ strings of $\theStringSet_s$ misses at most $\noiseLevel$ strings of $\theStringSet'$, so $\theLanguageCollection(\theStringSet', \noiseLevel) \supseteq \theLanguageCollection(\theStringSet_s, \noiseLevel) \ni \theLanguage$. Therefore, $\theLanguageCollection(\theStringSet', \noiseLevel)$ is non-empty and $\langle \theStringSet', \noiseLevel \rangle_\theLanguageCollection \subseteq X$ is finite.

\noindent\textit{Size:} Each $\theLanguage' \in \theLanguageCollection(\theStringSet', \noiseLevel)$ excludes at most $\noiseLevel$ strings of $\theStringSet'$, so at most $\noiseLevel|\theLanguageCollection|$ strings of $\theStringSet'$ lie outside $\langle \theStringSet', \noiseLevel \rangle_\theLanguageCollection$. So $|\langle \theStringSet', \noiseLevel \rangle_\theLanguageCollection| \geq \distinctNumExamples - \noiseLevel|\theLanguageCollection|$.

$(\Leftarrow)$ Let $\theCurrentStringSet = \theString_1, \ldots, \theString_\theTimestep$ be as in the statement. Since $\theLanguageCollection(\theCurrentStringSet, \noiseLevel)$ and $\langle \theCurrentStringSet, \noiseLevel \rangle_\theLanguageCollection$ depend only on the distinct strings of $\theCurrentStringSet$, we may assume these strings are distinct, so $\theTimestep = \distinctNumExamples$. Let $X = \langle \theCurrentStringSet, \noiseLevel \rangle_\theLanguageCollection$ and let $\learner$ be any generator. The adversary first presents $\theString_1, \ldots, \theString_\theTimestep$. For every $\theLanguage \in \theLanguageCollection(\theCurrentStringSet, \noiseLevel)$, at most $\noiseLevel$ of them lie outside $\theLanguage$, so this is a valid prefix of an enumeration of $\theLanguage$ at noise level $\noiseLevel$. The adversary then presents the strings of $X$ not in $\theCurrentStringSet$, one at a time. These are in every consistent language, so they are not noise for any of them, and they leave the consistent set unchanged.

At every time from $\theTimestep$ on, at least $\distinctNumExamples$ distinct strings have been seen. If at some such time $\learner$ outputs an already seen string, it fails. If it outputs a string outside $X$, some $\theLanguage \in \theLanguageCollection(\theCurrentStringSet, \noiseLevel)$ does not contain it; the adversary completes the stream with the remaining strings of $\theLanguage$, which is possible since $\theLanguage$ is infinite, giving an enumeration of $\theLanguage$ with at most $\noiseLevel$ noisy examples on which $\learner$ fails. Otherwise, $\learner$ always outputs an unseen string of $X$. Since $X$ is finite, all of $X$ has been presented after finitely many steps, after which every string of $X$ has been seen, so one of the previous two cases occurs.
\end{proof}

\ThmNoiseGeneric*

\begin{proof}
    Let $\distinctNumExamples \geq B(\theLanguageCollection) +
\noiseLevel|\theLanguageCollection|$. By
\cref{prop:noisy-unif-gen}, $\theLanguageCollection$ fails to be
$\distinctNumExamples$-generatable at noise level $\noiseLevel$ only if there is
an enumeration $\theCurrentStringSet$ with
$|\{\theString_1,\dots,\theString_\theTimestep\}| = \distinctNumExamples$ such that
$\theLanguageCollection(\theCurrentStringSet, \noiseLevel)$ is non-empty and
$\langle \theCurrentStringSet, \noiseLevel \rangle_{\theLanguageCollection}$ is
finite of size at least $\distinctNumExamples -
\noiseLevel|\theLanguageCollection|$. We show that no such enumeration exists.

Suppose $\theCurrentStringSet$ is an enumeration such that
$\theLanguageCollection(\theCurrentStringSet, \noiseLevel) \neq \emptyset$ and
$\langle \theCurrentStringSet, \noiseLevel \rangle_{\theLanguageCollection}$ is
finite. Write $\theLanguageCollSubset =
\theLanguageCollection(\theCurrentStringSet, \noiseLevel)$. By
\cref{def:noisy-closure},
\begin{equation}
    \langle \theCurrentStringSet, \noiseLevel \rangle_{\theLanguageCollection}
  = \bigcap_{\theLanguage \in \theLanguageCollSubset} \theLanguage ,
\end{equation}
and $\theLanguageCollSubset$ is a non-empty subset of
$\theLanguageCollection$. The noisy closure is therefore a finite intersection
of languages of $\theLanguageCollection$, which is exactly what is bounded in
the noiseless setting. Since $\theLanguageCollection$ is $B(\theLanguageCollection)$-generatable, by \cref{prop:arenas-prop-two-three}, every finite intersection of a non-empty subcollection of $\theLanguageCollection$ has size less than $B(\theLanguageCollection)$. So
\begin{equation}
    |\langle \theCurrentStringSet, \noiseLevel
  \rangle_{\theLanguageCollection}| < B(\theLanguageCollection)
  \leq \distinctNumExamples - \noiseLevel|\theLanguageCollection| ,
\end{equation}
where the second inequality holds because of the choice of $\distinctNumExamples$. The size
condition from \cref{prop:noisy-unif-gen} therefore fails for every enumeration, and
$\theLanguageCollection$ is $\distinctNumExamples$-generatable at noise level $\noiseLevel$.
\end{proof}

\subsection{Uncomputable bound for context-free languages}

\cite{arenas2026languagegenerationcomplexitybarriers} use the following lemma to establish that no computable bound on the sample complexity exists for context-free languages. Their proof reduces from the halting problem via a construction that encodes the computation of a Turing machine into the intersection of two CFGs. 

\begin{lemma}[Lemma 3.2, \citealp{arenas2026languagegenerationcomplexitybarriers}]\label{lemma:tm-construction}
    There exists an algorithm that, given a Turing Machine $\mathcal{M}$ with empty input, constructs two CFGs $\theGrammar_1$ and $\theGrammar_2$ such that, if $M$ halts in exactly $t$ computational steps, then 
    \begin{equation}
        |L(\theGrammar_1)\cap L(\theGrammar_2)|=2^t.
    \end{equation}
\end{lemma}

\begin{restatable}{corollary}{CFLNoisyGenThm}
    For any noise level $\noiseLevel$, there is no algorithm that, given two infinite CFGs $\theGrammar_1$ and $\theGrammar_2$, outputs $\distinctNumExamples\in\mathbb{N}$ such that $\{L(\theGrammar_1), L(\theGrammar_2)\}$ is uniformly $\distinctNumExamples$-generatable at noise level $\noiseLevel$. The algorithm may diverge if either language is finite.
\end{restatable}

\begin{proof}
    Fix some noise level $\noiseLevel$ and, for contradiction, assume there exists an algorithm such that, given two infinite CFGs $G_1$ and $G_2$, outputs $\distinctNumExamples$ such that the collection $\{L(G_1),L(G_2)\}$ is $\distinctNumExamples$-generatable at noise level $\noiseLevel$. Given a Turing Machine $M$, we construct CFGs $G_1$ and $G_2$ as in \cref{lemma:tm-construction}. We know that the finiteness of context-free languages is decidable. We distinguish three cases: 

    \noindent \textit{$L(G_1)$ is finite:} Choose $t$ such that $2^t > |L(G_1)|\geq |L(G_1)\cap L(G_2)|$. Then $M$ cannot halt in $t$ or more steps. Simulate $M$ for $t-1$ steps and conclude that it does not halt if it does not reach a halting configuration.

    \noindent \textit{$L(G_2)$ is finite:} Analogous to previous case.

    \noindent \textit{Both $L(G_1)$ and $L(G_2)$ are infinite:} Apply the algorithm and get $\distinctNumExamples$ such that $\{L(G_1), L(G_2)\}$ is $\distinctNumExamples$-generatable at noise level $\noiseLevel$. Since $\distinctNumExamples$-generatability implies $\distinctNumExamples'$-generatability for every $\distinctNumExamples'\geq \distinctNumExamples$, we may assume $\distinctNumExamples\geq 2\noiseLevel$. Choose $t$ such that $2^t \geq \distinctNumExamples - 2\noiseLevel$. We claim that $M$ cannot halt in $s\geq t$ steps.

    Suppose $M$ halts in $s\geq t$ steps. Then $|L(G_1)\cap L(G_2)|=2^s$. Since both $L(G_1)$ and $L(G_2)$ are infinite and their intersection is finite, both $L(G_1)\setminus L(G_2)$ and $L(G_2)\setminus L(G_1)$ are infinite. We construct an enumeration $\theStringSet$ consisting of $\distinctNumExamples-2\noiseLevel$ distinct elements from $L(G_1)\cap L(G_2)$, $\noiseLevel$ elements from $L(G_1)\setminus L(G_2)$, and $\noiseLevel$ elements from $L(G_2)\setminus L(G_1)$, giving $|\theStringSet|= \distinctNumExamples$. We now verify the conditions of \cref{prop:noisy-unif-gen} with $\theStringSet$ and $\distinctNumExamples$.     
    
    \noindent Since $|\theStringSet \cap L(G_1)|=\distinctNumExamples-\noiseLevel$ and $|\theStringSet \cap L(G_2)|=\distinctNumExamples-\noiseLevel$, it holds that $\theLanguageCollection(\theStringSet,\noiseLevel)=\{L(G_1), L(G_2)\}$, which is non-empty. Furthermore, $\left<\theStringSet,\noiseLevel\right>_\theLanguageCollection= L(G_1)\cap L(G_2)$, which is finite of size $2^s$. Finally, we need $|\left<\theStringSet,\noiseLevel\right>_\theLanguageCollection| \geq \distinctNumExamples - \noiseLevel |\theLanguageCollection|$. Trivially, $2^s \geq 2^t \geq \distinctNumExamples -2 \noiseLevel$. Therefore, $\{L(G_1), L(G_2)\}$ is not $\distinctNumExamples$-generatable at noise level $\noiseLevel$, contradicting the output of the algorithm. Therefore, $M$ cannot halt in $s\geq t$ steps and it suffices to simulate it for $t-1$ steps to decide whether it halts. In all cases, we have decided the halting problem, therefore the algorithm cannot exist.
\end{proof}

\subsection{Bounds for regular languages}

\begin{restatable}{corollary}{RegNoisyGenThm}\label{thm-reg-noisy-gen}
    Let $\theLanguageCollection$ be a finite collection of infinite regular languages over an alphabet of size $\theAlphabetSize\geq 2$ such that each $\theLanguage\in\theLanguageCollection$ is accepted by an NFA with at most $s$ states. Then $\theLanguageCollection$ is $(\theAlphabetSize^{s^{|\theLanguageCollection|}}+\noiseLevel |\theLanguageCollection|)$-generatable at noise level $\noiseLevel$.
\end{restatable}

\begin{proof}
By Theorem~4.1 of \citet{arenas2026languagegenerationcomplexitybarriers}, $\theLanguageCollection$ is $\theAlphabetSize^{s^{|\theLanguageCollection|}}$-generatable. Combining this with \cref{thm:noise-generic}, we get the desired bound.
\end{proof} 

\begin{theorem}
    There exists $c\in\mathbb{N}$ such that for every $\noiseLevel, l,n\in\mathbb{N}$ with $l \geq 1$ and $n \geq 1$, there exists a collection $\theLanguageCollection$ of infinite binary regular languages such that $|\theLanguageCollection|=2^l$, each $\theLanguage\in\theLanguageCollection$ is accepted by a DFA with at most $c \cdot l \cdot n$ states, and $\theLanguageCollection$ is not $(2^{n^{|\theLanguageCollection|}}+\noiseLevel |\theLanguageCollection|)$-generatable at noise level $\noiseLevel$.
\end{theorem}

\begin{proof}
    We use the same collection $\theLanguageCollection=\{\theLanguage_{w}^n\mid w\in \{0,1\}^l\}$ as \cite{arenas2026languagegenerationcomplexitybarriers}. By their construction, each $\theLanguage_{w}^n$ is accepted by a DFA with at most $c \cdot l \cdot n$ states, each language is infinite and 
    \begin{equation}
        \left|\bigcap_{w \in \{0,1\}^l} \theLanguage_{w}^n\right| \geq 2^{n^{|\theLanguageCollection|}},
    \end{equation}
    and this intersection is finite. 

    We use \cref{prop:noisy-unif-gen} to prove that $\theLanguageCollection$ is not $(2^{n^{|\theLanguageCollection|}} + \noiseLevel \cdot 2^l)$-generatable at noise level $\noiseLevel$. Let $\distinctNumExamples=2^{n^{|\theLanguageCollection|}} + \noiseLevel \cdot 2^l$ and choose $\theStringSet$ such that it contains $2^{n^{|\theLanguageCollection|}}$ distinct elements from the intersection and, for each $w\in\{0,1\}^l$, $\noiseLevel$ elements from $\bigcap_{w'\neq w} \theLanguage_{w'}^n$. We can prove that the set $\bigcap_{w'\neq w} \theLanguage_{w'}^n$ is infinite, therefore we can always pick $\noiseLevel$ elements from it. To show this, consider a binary string $b\in\{0,1\}^l$ and a string of the form $s_b=w b w b \cdots w$, with $j\geq n+1$ odd blocks equal to $w$, and even blocks equal to $b$. This violates the condition of language $\theLanguage_w^n$, therefore $s_b \notin \theLanguage_w^n$. For every $w' \neq w$, the string $s_b$ has zero odd $w'$ blocks, so the condition of language $\theLanguage_{w'}^n$ is trivially satisfied. Because the length of such strings can become arbitrarily large, there are infinitely many strings.
    We now verify the conditions of \cref{prop:noisy-unif-gen}.\looseness=-1

    For each language $\theLanguage_w^n$, there are $\noiseLevel$ samples outside of it. 
    So for every $\theLanguage_w^n\in\theLanguageCollection$, $|\theStringSet\cap\theLanguage_w^n|=\distinctNumExamples - \noiseLevel$. Every language is consistent with $\theStringSet$ at noise level $\noiseLevel$, by construction. Therefore $\theLanguageCollection(\theStringSet,\noiseLevel)=\theLanguageCollection$.

    Since $\theLanguageCollection(\theStringSet,\noiseLevel)=\theLanguageCollection$, it holds that the noisy closure $\left<\theStringSet,\noiseLevel\right>_\theLanguageCollection=\bigcap_{\theLanguage\in\theLanguageCollection(\theStringSet,\noiseLevel)}\theLanguage=\bigcap_{w \in \{0,1\}^l} \theLanguage_{w}^n$ is finite by the proof of \cite{arenas2026languagegenerationcomplexitybarriers}.

    We need to show that $|\left<\theStringSet,\noiseLevel\right>_\theLanguageCollection|\geq \distinctNumExamples - \noiseLevel |\theLanguageCollection|$. The right-hand side equals $2^{n^{|\theLanguageCollection|}} + \noiseLevel |\theLanguageCollection| - \noiseLevel |\theLanguageCollection| = 2^{n^{|\theLanguageCollection|}}$, and $|\left<\theStringSet,\noiseLevel\right>_\theLanguageCollection| \geq 2^{n^{|\theLanguageCollection|}}$. So $\theLanguageCollection$ is not $\distinctNumExamples$-generatable at noise level $\noiseLevel$.
\end{proof}

\subsection{Bounds for LTT languages}

\begin{restatable}{corollary}{LTTNoisyGenThm}
    Let $\theLanguageCollection$ be a finite collection of infinite languages over an alphabet of size $\theAlphabetSize \geq 2$ such that each language $\theLanguage\in\theLanguageCollection$ is defined by an LTT of size at most $s$. Then $\theLanguageCollection$ is $(\theAlphabetSize^{3s |\theLanguageCollection|((s |\theLanguageCollection|)^2+1)^2}+\noiseLevel |\theLanguageCollection|)$-generatable at noise level $\noiseLevel$.
\end{restatable}

\begin{proof}
    By Theorem~5.1 of \citet{arenas2026languagegenerationcomplexitybarriers}, $\theLanguageCollection$ is $\theAlphabetSize^{3s|\theLanguageCollection|((s|\theLanguageCollection|)^2+1)^2}$-generatable. Combining this with \cref{thm:noise-generic}, we get the desired bound.
\end{proof}

\begin{theorem}
    
    For every $n \in \mathbb{N}$ with $n \geq 4$, there exists a collection $\theLanguageCollection$ consisting of two infinite languages over an alphabet of size $n$ such that each language $\theLanguage\in\theLanguageCollection$ is defined by a profile of size at most $7(n+1)$, and $\theLanguageCollection$ is not $(2^n+2\noiseLevel)$-generatable at noise level $\noiseLevel$.
\end{theorem}

\begin{proof}
    We use the same two languages $L(P_1)$ and $L(P_2)$ from \cite{arenas2026languagegenerationcomplexitybarriers}, defined over the alphabet $\theAlphabet=\{1,2,\ldots, n\}$, such that 
    \begin{align}
        P_1 &= (\varepsilon, \varepsilon, \{(i, \leq 1)\colon i \text{ is odd} \} \cup \{(i, \geq 1)\colon i \text{ is even} \}), \\
        P_2 &= (\varepsilon, \varepsilon, \{(i, \geq 1)\colon i \text{ is odd} \} \cup \{(i, \leq 1)\colon i \text{ is even} \}).
    \end{align}
    Both $L(P_1)$ and $L(P_2)$ are infinite, $|L(P_1) \cap L(P_2)|=n!$, and each profile has length at most $7(n+1)$. Since $n! \geq 2^n$ for $n\geq 4$, the intersection has size at least $2^n$. Since the languages are infinite and their intersection is finite, both $L(P_1)\setminus L(P_2)$ and $L(P_2)\setminus L(P_1)$ are infinite. Choose an enumeration $\theStringSet$ consisting of $\distinctNumExamples=2^n+2\noiseLevel$ distinct elements such that $2^n$ elements are chosen from $L(P_2)\cap L(P_1)$, $\noiseLevel$ elements are chosen from $L(P_1)\setminus L(P_2)$ and $\noiseLevel$ elements are chosen from $L(P_2)\setminus L(P_1)$. We now verify the conditions of \cref{prop:noisy-unif-gen}.

    There are exactly $\noiseLevel$ elements chosen from $L(P_2)\setminus L(P_1)$, and these are the only elements of $\theStringSet$ outside $L(P_1)$, therefore $|\theStringSet \cap L(P_1)|=\distinctNumExamples - \noiseLevel$. Similarly, $|\theStringSet \cap L(P_2)|=\distinctNumExamples - \noiseLevel$. Therefore, $\theLanguageCollection(\theStringSet, \noiseLevel)=\{L(P_1), L(P_2)\}$.

    Since $\theLanguageCollection(\theStringSet, \noiseLevel)=\{L(P_1), L(P_2)\}$, it holds that $\left<\theStringSet,\noiseLevel\right>_\theLanguageCollection= L(P_1)\cap L(P_2)$, which is finite of size $n!$.

    We need to show that $|\left<\theStringSet,\noiseLevel\right>_\theLanguageCollection|\geq \distinctNumExamples - \noiseLevel |\theLanguageCollection|$. It holds that $n! \geq 2^n + 2 \noiseLevel - 2 \noiseLevel=2^n$ for any $n \geq 4$.

    By \cref{prop:noisy-unif-gen}, $\theLanguageCollection$ is not $(2^n+2 \noiseLevel)$-generatable at noise level $\noiseLevel$.
\end{proof}

\subsection{Bounds for partially-ordered languages}

\begin{restatable}{corollary}{ThmPTNoisyGen}
    Let $\theLanguageCollection$ be a finite collection of infinite PO languages, over an alphabet of size $\theAlphabetSize \ge 2$, each recognized by a PONFA with at most $s$ states. Then $\theLanguageCollection$ is $(\theAlphabetSize^{|\theLanguageCollection|(s-1)+1}+\noiseLevel |\theLanguageCollection|)$-generatable at noise level $\noiseLevel$.\looseness=-1
\end{restatable}

\begin{proof}
By \cref{thm:lt-bound}, $\theLanguageCollection$ is $\theAlphabetSize^{|\theLanguageCollection|(s-1)+1}$-generatable. Combining this with \cref{thm:noise-generic}, we get the desired bound.
\end{proof}

\begin{theorem}
    For every integer $s\geq 2$ and every noise level
    $\noiseLevel\in\mathbb{N}$, there exists a collection
    $\theLanguageCollection$ of two infinite PO languages over
    a binary alphabet, each recognized by a PONFA with $s$ states,
    such that $\theLanguageCollection$ is not
    $\left(\binom{2s}{s}-1+2\noiseLevel\right)$-generatable
    at noise level $\noiseLevel$.
\end{theorem}

\begin{proof}
    We use the same two languages $\theLanguage_1$ and
    $\theLanguage_2$ from the noiseless lower bound. Their
    intersection is finite and has size
    $N=\binom{2s}{s}-1$. Both
    $\theLanguage_1\setminus\theLanguage_2$ and
    $\theLanguage_2\setminus\theLanguage_1$ are infinite:
    they contain $b^j$ and $a^j$, respectively, for every $j\geq s$.

    Construct a sequence $S$ of distinct strings by taking all
    $N$ elements of $\theLanguage_1\cap\theLanguage_2$,
    together with $\noiseLevel$ elements from each of
    $\theLanguage_1\setminus\theLanguage_2$ and
    $\theLanguage_2\setminus\theLanguage_1$.
    Thus, $|S|=N+2\noiseLevel$.
    Each language excludes exactly $\noiseLevel$ strings of $S$,
    so both remain consistent at noise level $\noiseLevel$.
    The noisy closure is therefore
    $\langle S,\noiseLevel\rangle_{\theLanguageCollection}
    =\theLanguage_1\cap\theLanguage_2$,
    which is finite and has size
    $N=|S|-\noiseLevel|\theLanguageCollection|$.
    By \cref{prop:noisy-unif-gen},
    $\theLanguageCollection$ is not
    $\left(\binom{2s}{s}-1+2\noiseLevel\right)$-generatable
    at noise level $\noiseLevel$.
\end{proof}

\section{Prompted uniform generation}\label{sec:prompted-gen-bounds-app}

We first adapt the results of \citet{raman-generation-plmr} to our setting. Note that \citet{raman-generation-plmr} translated the prompted setting of \citet{kleinberg-2024-language-generation} into their multiclass setting by labeling each continuation with its prompt. Since a hypothesis in their setting assigns a single prompt to each continuation, this translation requires every continuation to follow at most one prompt in each language. There is no such restriction in our setting: if $ab, cb \in \theLanguage$, for instance, the continuation $b$ is valid for both prompts $a$ and $c$.

\begin{definition}
    A language collection $\theLanguageCollection$ is \defn{prompted uniformly generatable} if there exist a generator $\learner$ and an $\distinctNumExamples \in \mathbb{N}$ such that for every prompt $\thePrompt \in \thePromptDomain$, every $\theTrueLanguage \in \theLanguageCollection$ and any enumeration $\theString_1, \theString_2, \ldots$ of $\theTrueLanguage$, if there exists a $\theTimestepStar \in \mathbb{N}$ such that $|\{\theString_1, \ldots, \theString_\theTimestepStar\}^\thePrompt| = \distinctNumExamples$, then for all $\theTimestep \geq \theTimestepStar$, $\learner$ outputs, on input $\theString_1, \ldots, \theString_\theTimestep$ and $\thePrompt$, a string $\theGeneratedString$ such that $\thePrompt \cdot \theGeneratedString \in \theTrueLanguage \setminus \{\theString_1, \ldots, \theString_\theTimestep\}$. In this case, we say that $\theLanguageCollection$ is \defn{$\distinctNumExamples$-prompted uniformly generatable}.
\end{definition}

Like \citet{raman-generation-plmr}, we only require that the learner generates correctly for a prompt $\thePrompt$ after seeing sufficiently many examples continuing $\thePrompt$. In this respect, our setting is weaker than the prompted generation in the limit of \citet{kleinberg-2024-language-generation}, which requires correct completions after some time for all prompts, including prompts without any previously seen continuation. On the other hand, we require a single number of examples that suffices for every target language, enumeration and prompt, while \citet{kleinberg-2024-language-generation} only require correct generation in the limit.

\begin{definition}\label{def:prompted-closure-dim}
    For a prompt $\thePrompt \in \thePromptDomain$, the \defn{prompted closure dimension} $\promptedClosureDim(\theLanguageCollection, \thePrompt)$ of a collection $\theLanguageCollection$ at $\thePrompt$ is the supremum over $d \in \mathbb{N}$ satisfying: there exist a language $\theTrueLanguage \in \theLanguageCollection$ and an enumeration $\theStringSet$ of $\theTrueLanguage$ such that there exists a timestep $\theTimestep$ with $|\theCurrentStringSet^\thePrompt| \geq d$ and $\left<\theCurrentStringSet, \thePrompt\right>_\theLanguageCollection \setminus \theCurrentStringSet^\thePrompt = \emptyset$.
The \defn{prompted closure dimension} of $\theLanguageCollection$ is $\promptedClosureDim(\theLanguageCollection) = \sup_{\thePrompt \in \thePromptDomain} \promptedClosureDim(\theLanguageCollection, \thePrompt)$.
\end{definition}

\begin{restatable}{theorem}{PromptedGenClosureThm}\label{thm:unif-prompted-generation}
    A collection $\theLanguageCollection$ is prompted uniformly generatable if and only if $\promptedClosureDim(\theLanguageCollection) < \infty$.
\end{restatable}

In the following, we adapt the proof of \citet{raman-generation-plmr} to our setting.

\begin{proof}
($\Rightarrow$) Suppose $\promptedClosureDim(\theLanguageCollection) = \infty$, and fix any generator $\learner$ and any $\distinctNumExamples \in \mathbb{N}$. Since $\sup_{\thePrompt} \promptedClosureDim(\theLanguageCollection, \thePrompt) = \infty$, there is a prompt $\thePrompt$ with $\promptedClosureDim(\theLanguageCollection, \thePrompt) \geq \distinctNumExamples$, so there exist $\theTrueLanguage \in \theLanguageCollection$, an enumeration $\theStringSet$ of $\theTrueLanguage$, and a timestep $\theTimestep$ with $|\theCurrentStringSet^\thePrompt| \geq \distinctNumExamples$ and $\langle \theCurrentStringSet, \thePrompt \rangle_\theLanguageCollection \subseteq \theCurrentStringSet^\thePrompt$. At time $\theTimestep$, at least $\distinctNumExamples$ distinct continuations of $\thePrompt$ have been seen. Let $a = \learner(\theCurrentStringSet, \thePrompt)$. If $a \in \theCurrentStringSet^\thePrompt$, then $\thePrompt \cdot a$ has already been seen, and $\learner$ fails on $\theTrueLanguage$. Otherwise $a \notin \langle \theCurrentStringSet, \thePrompt \rangle_\theLanguageCollection$, so some language $\theLanguage$ consistent with $\theCurrentStringSet$ satisfies $a \notin \theLanguage^\thePrompt$. Since the strings of $\theCurrentStringSet$ are in $\theLanguage$ and $\theLanguage$ is infinite, $\theCurrentStringSet$ extends to an enumeration of $\theLanguage$, on which $\learner$ fails at time $\theTimestep$. Since $\learner$ and $\distinctNumExamples$ were arbitrary, $\theLanguageCollection$ is not prompted uniformly generatable.

($\Leftarrow$) Suppose $d = \promptedClosureDim(\theLanguageCollection) < \infty$, and let $\distinctNumExamples = d + 1$. Given $\theCurrentStringSet$ and $\thePrompt$, let $\learner$ output an element of $\langle \theCurrentStringSet, \thePrompt \rangle_\theLanguageCollection \setminus \theCurrentStringSet^\thePrompt$ if this set is non-empty, and an arbitrary string otherwise. Fix $\thePrompt \in \thePromptDomain$, $\theTrueLanguage \in \theLanguageCollection$, an enumeration of $\theTrueLanguage$, and $\theTimestepStar$ with $|\theStringSet_{\theTimestepStar}^\thePrompt| = \distinctNumExamples$. For every $\theTimestep \geq \theTimestepStar$, we have $|\theCurrentStringSet^\thePrompt| \geq d + 1 > \promptedClosureDim(\theLanguageCollection, \thePrompt)$, so $\langle \theCurrentStringSet, \thePrompt \rangle_\theLanguageCollection \setminus \theCurrentStringSet^\thePrompt \neq \emptyset$; otherwise $\theTrueLanguage$, its enumeration and $\theTimestep$ would witness $\promptedClosureDim(\theLanguageCollection, \thePrompt) \geq d + 1$. Since $\theTrueLanguage$ is consistent with $\theCurrentStringSet$, the closure is contained in $\theTrueLanguage^\thePrompt$, so $\learner$ outputs an $a$ with $\thePrompt \cdot a \in \theTrueLanguage \setminus \theCurrentStringSet$.
\end{proof}

\PromptedGenFiniteCollProp*

\begin{proof}
It suffices to show that $\promptedClosureDim(\theLanguageCollection)<\infty$. Since both $\theLanguageCollection$ and $\thePromptDomain$ are finite, the set 
\begin{equation}
    \mathcal{I}= \left\{\bigcap_{\theLanguage\in\theLanguageCollSubset}\theLanguage^\thePrompt\colon \theLanguageCollSubset\subseteq \theLanguageCollection, \thePrompt\in\thePromptDomain, \theLanguageCollSubset\neq \emptyset\right\}
\end{equation}
is finite. Let $D$ be the maximum size of any element of $\mathcal{I}$ that is finite, or $0$ if there is no such element. 

We prove that $\promptedClosureDim(\theLanguageCollection)\leq D$. Assume for contradiction that $\promptedClosureDim(\theLanguageCollection)> D$. Then there exist $\theTrueLanguage\in\theLanguageCollection$, $\thePrompt\in\thePromptDomain$, and an enumeration $\theStringSet$ of $\theTrueLanguage$ with a timestep $\theTimestep$ such that $|\theCurrentStringSet^\thePrompt|\geq D+1$, and
\begin{equation}
    \left(\bigcap_{\theLanguage\in\theLanguageCollection(\theCurrentStringSet,\thePrompt)}\theLanguage^\thePrompt\right)\setminus \theCurrentStringSet^\thePrompt = \emptyset.
\end{equation}
Let $X=\bigcap_{\theLanguage\in\theLanguageCollection(\theCurrentStringSet,\thePrompt)}\theLanguage^\thePrompt$. Since each $\theLanguage\in\theLanguageCollection(\theCurrentStringSet,\thePrompt)$ contains $\theCurrentStringSet$, we have $\theCurrentStringSet^\thePrompt\subseteq \theLanguage^\thePrompt$, thus $\theCurrentStringSet^\thePrompt\subseteq X$. By combining this with $X \subseteq \theCurrentStringSet^\thePrompt$ from the failure condition, we get $X=\theCurrentStringSet^\thePrompt$. 

Since $|\theCurrentStringSet^\thePrompt|\geq D+1$, we have $|X|\geq D+1$. But $X \in \mathcal{I}$ and $X$ is finite, so $|X|\leq D$ by the definition of $D$. This leads to a contradiction.
\end{proof}

In general, the prompt domain $\thePromptDomain$ being finite is essential. The following example shows a finite collection consisting of only two infinite languages that is not prompted uniformly generatable. This example is adapted from \cite{raman-generation-plmr} to match our prompted generation setting.

\begin{example}
Let $\Sigma=\{1,\#,a\}$ and $\thePromptDomain=\{1\#, 11\#, 111\#,\ldots\}$. We use $p_n$ to denote the prompt $1^n\#$. Note that no prompt is a prefix of another.

For each prompt $p_n$, let $A_n=\{a^{n^2+1},\ldots,a^{n^2+n}\}$ be a set of $n$ suffixes. Notice that the sets $A_1, A_2,\ldots$ are pairwise disjoint.

Now define the two languages
\begin{align}
\theLanguage_1 &= \{p_n\cdot x: n\in\mathbb{N}, x\in A_n\}\cup\{p_n\cdot a^k: n\in\mathbb{N}, k>n^2+n, k \text{ even}\},\\
\theLanguage_2 &= \{p_n\cdot x: n\in\mathbb{N}, x\in A_n\}\cup\{p_n\cdot a^k: n\in\mathbb{N}, k>n^2+n, k \text{ odd}\}.
\end{align}
Both languages are infinite, and for every prompt $p_n$, both projections $\theLanguage_1^{p_n}$ and $\theLanguage_2^{p_n}$ are infinite. Every string of $\theLanguage_1$ or $\theLanguage_2$ beginning with $p_n$ is of the form $p_n\cdot y$ for a unique $n$, so $\theLanguage_1^{p_n}\cap\theLanguage_2^{p_n}=A_n$, which is finite of size $n$.
\end{example}

\PropPromptedUnifGen*

\begin{proof}
($\Rightarrow$) Suppose $\theLanguageCollection$ is not $\distinctNumExamples$-prompted uniformly generatable. Let $\learner^*$ output an unseen element of $\langle \theCurrentStringSet, \thePrompt \rangle_\theLanguageCollection$ whenever one exists. Since $\learner^*$ fails, there exist a prompt $\thePrompt$, a language $\theTrueLanguage \in \theLanguageCollection$, an enumeration of $\theTrueLanguage$, and a timestep $\theTimestep$ with $|\theCurrentStringSet^\thePrompt| \geq \distinctNumExamples$ such that $\learner^*(\theCurrentStringSet, \thePrompt) \notin \theTrueLanguage^\thePrompt \setminus \theCurrentStringSet^\thePrompt$. Let $\theLanguageCollSubset$ be the set of languages consistent with $\theCurrentStringSet$, and $X = \bigcap_{\theLanguage \in \theLanguageCollSubset} \theLanguage^\thePrompt = \langle \theCurrentStringSet, \thePrompt \rangle_\theLanguageCollection$. Then $\theTrueLanguage \in \theLanguageCollSubset$, so $\theLanguageCollSubset$ is non-empty and $X \subseteq \theTrueLanguage^\thePrompt$.

\noindent\textit{$X$ is finite:} Otherwise, $X$ contains an unseen continuation, which $\learner^*$ outputs and which lies in $\theTrueLanguage^\thePrompt$, contradicting that $\learner^*$ fails.

\noindent\textit{$|X| \geq \distinctNumExamples$:} Every $\theLanguage \in \theLanguageCollSubset$ contains the strings of $\theCurrentStringSet$, so $\theCurrentStringSet^\thePrompt \subseteq \theLanguage^\thePrompt$, hence $\theCurrentStringSet^\thePrompt \subseteq X$ and $|X| \geq |\theCurrentStringSet^\thePrompt| \geq \distinctNumExamples$.

($\Leftarrow$) Let $\thePrompt$ and $\theLanguageCollSubset$ be as in the statement, $X = \bigcap_{\theLanguage \in \theLanguageCollSubset} \theLanguage^\thePrompt$, and let $\learner$ be any generator. The adversary presents the strings $\thePrompt \cdot x$ for $x \in X$, one at a time. These lie in every $\theLanguage \in \theLanguageCollSubset$, so at every time, the strings presented so far form a valid prefix of an enumeration of every $\theLanguage \in \theLanguageCollSubset$. Consider any time after the $\distinctNumExamples$-th string has been presented. At that point, at least $\distinctNumExamples$ distinct continuations of $\thePrompt$ have been seen. If $\learner$ outputs a continuation already seen, it fails. If it outputs $a \notin X$, some $\theLanguage \in \theLanguageCollSubset$ satisfies $a \notin \theLanguage^\thePrompt$. The adversary completes the enumeration with the remaining strings of $\theLanguage$, which is possible since $\theLanguage$ is infinite, and $\learner$ fails on $\theLanguage$. Otherwise $\learner$ outputs an unseen element of $X$. Since $X$ is finite, all of $X$ has been presented after finitely many steps, after which one of the previous two cases occurs.
\end{proof}

\ThmPromptedGeneric*

\begin{proof}
For $\thePrompt \in \thePromptDomain$, write $\theLanguageCollection^\thePrompt = \{\theLanguage^\thePrompt \colon \theLanguage \in \theLanguageCollection\}$. By Assumption~1, $\theLanguageCollection^\thePrompt$ is a finite collection of infinite languages, and for each class we study the representation is effectively closed under left quotient. The recipe of \cref{sec:recipe} therefore applies to $\theLanguageCollection^\thePrompt$: every finite intersection of a non-empty subcollection of $\theLanguageCollection^\thePrompt$ has size less than $B(\theLanguageCollection^\thePrompt)$.

Let $\distinctNumExamples = \sup_\thePrompt B(\theLanguageCollection^\thePrompt) < \infty$, and suppose, for contradiction, that $\theLanguageCollection$ is not $\distinctNumExamples$-prompted uniformly generatable. By \cref{prop:prompted-closure-dim}, there exist a prompt $\thePrompt$ and a non-empty $\theLanguageCollSubset \subseteq \theLanguageCollection$ such that $\bigcap_{\theLanguage \in \theLanguageCollSubset} \theLanguage^\thePrompt$ is finite and has size at least $\distinctNumExamples \geq B(\theLanguageCollection^\thePrompt)$. But $\{\theLanguage^\thePrompt \colon \theLanguage \in \theLanguageCollSubset\}$ is a non-empty subcollection of $\theLanguageCollection^\thePrompt$, so this finite intersection has size less than $B(\theLanguageCollection^\thePrompt)$, which gives a contradiction.
\end{proof}

\subsection{Computing the left quotient}

\paragraph{Regular languages.} Given an NFA $\theAutomaton=\theFSATuple$ recognizing $\theLanguage$, the left quotient $\theLanguage^\thePrompt$ is recognized by the NFA $\theAutomaton'=(\theAlphabet,\theStates,\theInitStates_\thePrompt,\theFinalStates,\theTransitions)$, where $\theInitStates_\thePrompt=\{\theState\in\theStates\colon \exists \theInitState\in\theInitStates \text{ s.t. there is a path from } \theInitState \text{ to }\theState \text{ reading }\thePrompt\}$. Some states can become unreachable and trimming can remove them, but in general the number of states is unchanged.

\paragraph{LTT languages.} It suffices to consider a single profile $P = (Pr, Su, In)$ of size $s$. Let $\lambda = \max\{|w| : (w, \theta t) \in In\}$, or $\lambda = 1$ if $In = \emptyset$. Note that $\lambda \leq s$. We distinguish strings $\theString$ by their length.
 Consider strings $\theString$ with $|\theString| \geq \max(\lambda - 1, |Su|)$. Since $|\theString| \geq |Su|$, $Su$ is a suffix of $\thePrompt \cdot \theString$ iff it is a suffix of $\theString$, so the suffix is unchanged. For the prefix, if neither $\thePrompt$ nor $Pr$ is a prefix of the other, no string of $\theLanguage(P)$ begins with $\thePrompt$, so $\theLanguage(P)^\thePrompt = \emptyset$. Otherwise, let $Pr'$ be the portion of $Pr$ beyond $\thePrompt$ if $\thePrompt$ is a prefix of $Pr$, and $Pr' = \varepsilon$ otherwise.
Each occurrence of a word $w$ with $(w, \theta t) \in In$ in $\thePrompt \cdot \theString$ either lies entirely in $\thePrompt$, lies entirely in $\theString$, or straddles the boundary. The number $o_w$ of occurrences in $\thePrompt$ depends only on $w$ and $\thePrompt$. A straddling occurrence ends within the first $|w| - 1 \leq \lambda - 1$ symbols of $\theString$, so the number $c_w(y)$ of straddling occurrences is determined by $\thePrompt$ and the prefix $y$ of $\theString$ of length $\lambda - 1$.
For each $y \in \Sigma^{\lambda - 1}$ compatible with $Pr'$ (i.e., one is a prefix of the other), we therefore obtain a profile $P_y$ whose prefix is the longer of $Pr'$ and $y$, whose suffix is $Su$, and in which each constraint $(w, \theta t)$ is replaced by $(w, \theta t')$ with $t' = t - o_w - c_w(y)$. If $t' < 0$, the constraint is dropped when $\theta \in \{>, \geq\}$, since it is always satisfied, and $P_y$ is omitted when $\theta \in \{<, \leq, =\}$, since it is unsatisfiable.
Constraints are only dropped and thresholds only decrease, so $P_y$ differs in size from $P$ only through its prefix. Since $|Pr'| \leq |Pr|$ and $|y| = \lambda - 1$, this gives $|P_y| \leq s + \lambda - 1 < 2s$. There are at most $|\Sigma|^{\lambda - 1}$ such profiles.
The remaining strings of $\theLanguage(P)^\thePrompt$ have length less than $\max(\lambda - 1, |Su|) \leq s$, so they form a finite set $F_P$.
Hence $\theLanguage(P)^\thePrompt = \bigcup_y \theLanguage(P_y) \cup F_P$, a finite union of profiles of size less than $2s$ together with a finite set of strings of length less than $s$.

\paragraph{PO languages.} Given a PONFA $\theAutomaton$ recognizing $\theLanguage$, the left quotient $\theLanguage^\thePrompt$ is recognized by the automaton obtained from $\theAutomaton$ by shifting the initial states to $\theInitStates_\thePrompt$, just like for regular languages. Partial-orderedness is preserved by the shift. The number of states is unchanged in general.

\paragraph{Context-free languages.} Given a CFG generating $\theLanguage$, the left quotient $\theLanguage^\thePrompt$ is also context-free. However, the uncomputability barrier transfers.

\subsection{Sample complexity bounds}

The corollaries below follow from \cref{thm:prompt-generic} and the quotient constructions above. They hold for any prompt domain $\thePromptDomain$, finite or infinite.

\begin{corollary}
    Let $\theLanguageCollection$ be a finite collection of infinite regular languages over an alphabet of size $\theAlphabetSize\geq 2$ satisfying \cref{assumption:infinite-prompted-set}, each recognized by an NFA with at most $s$ states. Then $\theLanguageCollection$ is $\theAlphabetSize^{s^{|\theLanguageCollection|}}$-prompted uniformly generatable.
\end{corollary}
\begin{proof}
    For every prompt $\thePrompt$, each $\theLanguage^\thePrompt$ is recognized by an NFA with at most $s$ states, so $B(\{\theLanguage^\thePrompt\colon\theLanguage\in\theLanguageCollection\})=\theAlphabetSize^{s^{|\theLanguageCollection|}}$ independently of $\thePrompt$. The claim follows from \cref{thm:prompt-generic}.
\end{proof}

\begin{corollary}
    Let $\theLanguageCollection$ be a finite collection of infinite LTT languages over an alphabet of size $\theAlphabetSize\geq 2$ satisfying \cref{assumption:infinite-prompted-set}, each defined by profiles of size at most $s$. Then $\theLanguageCollection$ is $\theAlphabetSize^{3r(r^2+1)^2}$-prompted uniformly generatable, where $r=2s|\theLanguageCollection|$.
\end{corollary}
\begin{proof}
    For every prompt $\thePrompt$, each $\theLanguage^\thePrompt$ is a finite union of profiles of size less than $2s$ together with a finite set of strings of length less than $s$. A finite intersection of such quotients is a union of intersections of one such piece from each language. If all pieces are profiles, their intersection is a profile of size less than $r$, whose strings have length less than $3r(r^2+1)^2$ when it is finite. Otherwise, its strings have length less than $s$. Hence $B(\{\theLanguage^\thePrompt\colon\theLanguage\in\theLanguageCollection\})\leq\theAlphabetSize^{3r(r^2+1)^2}$ independently of $\thePrompt$, and the claim follows from \cref{thm:prompt-generic}.
\end{proof}

\begin{corollary}
    For any prompt $\thePrompt$, there is no algorithm that, given CFGs for two infinite languages $\theLanguage_1$ and $\theLanguage_2$ satisfying \cref{assumption:infinite-prompted-set} for the prompt domain $\thePromptDomain=\{\thePrompt\}$, always outputs $\distinctNumExamples$ such that $\{\theLanguage_1,\theLanguage_2\}$ is $\distinctNumExamples$-prompted uniformly generatable.
\end{corollary}
\begin{proof}
    Suppose such an algorithm exists. Given CFGs $\theGrammar_1,\theGrammar_2$ for two infinite languages, construct CFGs for $\thePrompt\cdot L(\theGrammar_1)$ and $\thePrompt\cdot L(\theGrammar_2)$ by adding a new start symbol $S'$ with the rule $S'\rightarrow \thePrompt\, S$. Their left quotients by $\thePrompt$ are $L(\theGrammar_1)$ and $L(\theGrammar_2)$, which are infinite, so \cref{assumption:infinite-prompted-set} holds for $\thePromptDomain=\{\thePrompt\}$. Every string of these languages begins with $\thePrompt$, so the number of distinct examples continuing $\thePrompt$ equals the number of distinct examples, and a continuation $a$ satisfies $\thePrompt\cdot a\in \thePrompt\cdot L(\theGrammar_i)$ if and only if $a\in L(\theGrammar_i)$. Hence $\{\thePrompt\cdot L(\theGrammar_1),\thePrompt\cdot L(\theGrammar_2)\}$ is $\distinctNumExamples$-prompted uniformly generatable if and only if $\{L(\theGrammar_1),L(\theGrammar_2)\}$ is $\distinctNumExamples$-generatable. The algorithm would therefore compute a valid bound for unprompted generation of two infinite context-free languages, which is impossible by \citet{arenas2026languagegenerationcomplexitybarriers}.
\end{proof}

\begin{corollary}
    Let $\theLanguageCollection$ be a finite collection of infinite PO languages over an alphabet of size $\theAlphabetSize\geq 2$ satisfying \cref{assumption:infinite-prompted-set}, each recognized by a PONFA with at most $s$ states. Then $\theLanguageCollection$ is $\theAlphabetSize^{|\theLanguageCollection|(s-1)+1}$-prompted uniformly generatable.
\end{corollary}
\begin{proof}
    For every prompt $\thePrompt$, each $\theLanguage^\thePrompt$ is recognized by a PONFA with at most $s$ states, so by \cref{thm:lt-bound}, $B(\{\theLanguage^\thePrompt\colon\theLanguage\in\theLanguageCollection\})=\theAlphabetSize^{|\theLanguageCollection|(s-1)+1}$ independently of $\thePrompt$. The claim follows from \cref{thm:prompt-generic}.
\end{proof}

\section{Uniform generation with feedback}\label{sec:feedback-gen-bounds-app}

\begin{definition}
    Let $\theNumQueries \in \mathbb{N}$. A language collection $\theLanguageCollection$ is \defn{uniformly $\distinctNumExamples$-generatable with $\theNumQueries$ feedback queries} if there exists a generator $\learner$ such that the following holds. At each timestep $\theTimestep$, after receiving $\theString_\theTimestep$ and before producing its output, $\learner$ may ask any number of queries of whether a string belongs to the target language and receive their answers, and it asks at most $\theNumQueries$ queries in total. For every $\theLanguage \in \theLanguageCollection$ and any enumeration $\theString_1, \theString_2, \ldots$ of $\theLanguage$, if there exists $\theTimestepStar \in \mathbb{N}$ such that $|\{\theString_1, \ldots, \theString_\theTimestepStar\}| = \distinctNumExamples$, then for all $\theTimestep \geq \theTimestepStar$, $\learner$ outputs a string in $\theLanguage \setminus \{\theString_1, \ldots, \theString_\theTimestep\}$. We say $\theLanguageCollection$ is \defn{uniformly generatable with $\theNumQueries$ feedback queries} if it is uniformly $\distinctNumExamples$-generatable with $\theNumQueries$ feedback queries for some $\distinctNumExamples$.
\end{definition}

We adapt the definition of feedback closure dimension of \cite{charikar2024exploringfacetslanguagegeneration}, which they call feedback-generation closure, to our setup. 
    
\begin{definition}\label{def:feedback-closure-dim}
    The \defn{feedback-augumented closure dimension} $\feedbackClosureDim(\theLanguageCollection, \theNumQueries)$ of a collection $\theLanguageCollection$ at query budget $\theNumQueries$ is the supremum over $d \in \mathbb{N}$ satisfying: for every generator $\learner$ that asks at most $\theNumQueries$ feedback queries, there exist a language $\theTrueLanguage \in \theLanguageCollection$, an enumeration $\theStringSet$ of $\theTrueLanguage$, and a timestep $\theTimestep$ with $|\{\theString_1, \ldots, \theString_\theTimestep\}| \geq d$ and
\begin{equation}
    \langle \theCurrentStringSet, Q_k \rangle_\theLanguageCollection \setminus \{\theString_1, \ldots, \theString_\theTimestep\} = \emptyset,
\end{equation}
where $Q_k$ denotes the $k \leq \theNumQueries$ queries asked by $\learner$ up to and including timestep $\theTimestep$ on this enumeration, answered according to $\theTrueLanguage$.
\end{definition}

\cite{charikar2024exploringfacetslanguagegeneration} give a characterization of the collections that are uniformly generatable with feedback. However, their setup is different, therefore we give the following analogous characterization.

\begin{restatable}{theorem}{FeedbackGenClosureCharacterization}\label{thm:unif-generation-feedback}
   Let $\theNumQueries \in \mathbb{N}$ be the feedback budget. A collection $\theLanguageCollection$ is uniformly generatable with $\theNumQueries$ feedback queries if and only if $\feedbackClosureDim(\theLanguageCollection, \theNumQueries) < \infty$.
\end{restatable}

The proof follows that of Lemmas~7.2 and~7.3 of \citet{charikar2024exploringfacetslanguagegeneration}, adapted to a fixed query budget.

\begin{proof}
($\Rightarrow$) Suppose $\feedbackClosureDim(\theLanguageCollection, \theNumQueries) = \infty$, and fix any generator $\learner$ asking at most $\theNumQueries$ queries and any $\distinctNumExamples \in \mathbb{N}$. By \cref{def:feedback-closure-dim}, applied with some $d \geq \distinctNumExamples$, there exist a target $\theTrueLanguage$, an enumeration of $\theTrueLanguage$, and a timestep $\theTimestep$ with $|\{\theString_1, \ldots, \theString_\theTimestep\}| \geq \distinctNumExamples$ such that $\langle \theCurrentStringSet, Q_k \rangle_\theLanguageCollection \setminus \{\theString_1, \ldots, \theString_\theTimestep\} = \emptyset$, where $Q_k$ are the queries asked by $\learner$ up to and including timestep $\theTimestep$, answered according to $\theTrueLanguage$. Let $a$ be the output of $\learner$ at timestep $\theTimestep$. If $a$ has already been seen, $\learner$ fails. Otherwise $a$ lies outside the closure, so some language $\theLanguage$ consistent with the examples and answers up to timestep $\theTimestep$ satisfies $a \notin \theLanguage$. The run up to timestep $\theTimestep$ is a valid prefix for target $\theLanguage$: all examples lie in $\theLanguage$, and all answers agree with $\theLanguage$. Completing the enumeration with the remaining strings of $\theLanguage$, which is possible since $\theLanguage$ is infinite, and answering later queries according to $\theLanguage$, $\learner$ still outputs $a \notin \theLanguage$ at timestep $\theTimestep$, after at least $\distinctNumExamples$ distinct strings. Since $\learner$ and $\distinctNumExamples$ were arbitrary, $\theLanguageCollection$ is not uniformly generatable with $\theNumQueries$ feedback queries.

($\Leftarrow$) Suppose $d = \feedbackClosureDim(\theLanguageCollection, \theNumQueries) < \infty$. The condition of \cref{def:feedback-closure-dim} fails for $d + 1$, so there is a generator $\learner$ asking at most $\theNumQueries$ queries such that for every target $\theTrueLanguage$, every enumeration of $\theTrueLanguage$, and every timestep $\theTimestep$ with at least $d + 1$ distinct strings, the closure $\langle \theCurrentStringSet, Q_k \rangle_\theLanguageCollection$ formed with the queries of $\learner$ contains an unseen string. Let $\learner'$ ask exactly the queries of $\learner$ and output an unseen string of this closure whenever one exists. Since the target is consistent with the examples and answers, the closure is contained in $\theTrueLanguage$, so from the time $d + 1$ distinct strings have been seen on, $\learner'$ outputs strings of $\theTrueLanguage$ not seen before. Hence $\theLanguageCollection$ is $(d+1)$-generatable with $\theNumQueries$ feedback queries.
\end{proof}

\begin{proposition}
Any finite collection $\theLanguageCollection$ is uniformly generatable with feedback.
\end{proposition}

\begin{proof}
Let $\theLanguageCollection$ be a finite collection such that $|\theLanguageCollection| = q$. It suffices to show that $\feedbackClosureDim(\theLanguageCollection, \theNumQueries) < \infty$ for all $\theNumQueries \in \mathbb{N}$.

By Theorem~36 of \citet{raman-generation-plmr}, all finite collections are uniformly generatable. Therefore, it holds that $\mathrm{C}(\theLanguageCollection) = \feedbackClosureDim(\theLanguageCollection, 0) < \infty$. With no feedback queries, the feedback closure reduces to the closure dimension of \citet{raman-generation-plmr}.

Fix any $\theNumQueries \in \mathbb{N}$. Since more queries shrink the consistent set, i.e., $\theLanguageCollection(\theCurrentStringSet, Q_\theNumQueries) \subseteq \theLanguageCollection(\theCurrentStringSet, \emptyset)$, the intersection $\bigcap_{\theLanguage \in \theLanguageCollection(\theCurrentStringSet, Q_\theNumQueries)} \theLanguage$ is larger, making it harder for the adversary to force $\left(\bigcap_{\theLanguage \in \theLanguageCollection(\theCurrentStringSet, Q_\theNumQueries)} \theLanguage\right) \setminus \theCurrentStringSet = \emptyset$. Therefore, $\feedbackClosureDim(\theLanguageCollection, \theNumQueries) \leq \feedbackClosureDim(\theLanguageCollection, 0) < \infty$ for all $\theNumQueries \in \mathbb{N}$.
\end{proof}

\PropUnifGenFeedback*

\begin{proof}
    $(\Rightarrow)$ Suppose $\theLanguageCollection$ is not $\distinctNumExamples$-generatable with $\theNumQueries$ feedback queries. Let $\learner$ be any generator asking at most $\theNumQueries$ feedback queries, and let $\learner^*$ be the generator that asks exactly the queries $\learner$ would ask on the same examples and answers, and outputs an arbitrary unseen element of the feedback-augmented closure whenever one exists. Since $\learner^*$ asks at most $\theNumQueries$ queries, it fails: there exist a language $\theTrueLanguage \in \theLanguageCollection$, an enumeration $\theStringSet = \theString_1, \theString_2, \ldots$ of $\theTrueLanguage$, and a timestep $\theTimestep$ with $|\{\theString_1, \ldots, \theString_\theTimestep\}| \geq \distinctNumExamples$ such that the output of $\learner^*$ at timestep $\theTimestep$ is not in $\theTrueLanguage \setminus \{\theString_1, \ldots, \theString_\theTimestep\}$. Let $Q_k$ be the queries asked up to and including timestep $\theTimestep$, which are the queries $\learner$ asks on this run, answered according to $\theTrueLanguage$, and let $X = \left<\theCurrentStringSet, Q_k\right>_\theLanguageCollection$. Since the examples lie in $\theTrueLanguage$ and the answers agree with $\theTrueLanguage$, we have $\theTrueLanguage \in \theLanguageCollection(\theCurrentStringSet, Q_k)$, so $X \neq \bot$ and $X \subseteq \theTrueLanguage$.

\textit{$X$ contains only strings of $\theCurrentStringSet$:} Otherwise $X$ contains an unseen string, which $\learner^*$ outputs at timestep $\theTimestep$, and which lies in $\theTrueLanguage$, contradicting that $\learner^*$ fails at timestep $\theTimestep$. In particular, $X$ is finite.

\textit{$|X| \geq \distinctNumExamples$:} Every $\theLanguage \in \theLanguageCollection(\theCurrentStringSet, Q_k)$ contains $\theString_1, \ldots, \theString_\theTimestep$, hence $\{\theString_1, \ldots, \theString_\theTimestep\} \subseteq X$ and $|X| \geq \distinctNumExamples$.

$(\Leftarrow)$ Let $\learner$ be any generator asking at most $\theNumQueries$ feedback queries. By hypothesis, there exist $\theTrueLanguage \in \theLanguageCollection$, an enumeration of $\theTrueLanguage$, and a timestep $\theTimestep$ with $|\{\theString_1, \ldots, \theString_\theTimestep\}| \geq \distinctNumExamples$ such that, for the queries $Q_k$ asked by $\learner$ up to and including timestep $\theTimestep$, answered according to $\theTrueLanguage$, the closure $X = \left<\theCurrentStringSet, Q_k\right>_\theLanguageCollection$ is finite, has size at least $\distinctNumExamples$, and contains only strings of $\theCurrentStringSet$. Let $a$ be the output of $\learner$ at timestep $\theTimestep$. We distinguish two cases:

\textit{$a \in \{\theString_1, \ldots, \theString_\theTimestep\}$:} Then $\learner$ outputs a string that has already been seen, and fails on $\theTrueLanguage$.

\textit{$a \notin \{\theString_1, \ldots, \theString_\theTimestep\}$:} Since $X$ contains only strings of $\theCurrentStringSet$, we have $a \notin X$, so there exists some $\theLanguage \in \theLanguageCollection(\theCurrentStringSet, Q_k)$ with $a \notin \theLanguage$. The run up to timestep $\theTimestep$ is a valid prefix for target $\theLanguage$: all examples lie in $\theLanguage$ and all answers agree with $\theLanguage$. The adversary completes the enumeration with the remaining strings of $\theLanguage$, which is possible since $\theLanguage$ is infinite, and answers later queries according to $\theLanguage$. On this run, $\learner$ still outputs $a \notin \theLanguage$ at timestep $\theTimestep$, after at least $\distinctNumExamples$ distinct strings, so $\learner$ fails on $\theLanguage$.

Since $\learner$ was arbitrary, $\theLanguageCollection$ is not $\distinctNumExamples$-generatable with $\theNumQueries$ feedback queries.
\end{proof}

\ThmFeedbackSimple*

\begin{proof}
    Let $\distinctNumExamples \geq \max\{B(\theLanguageCollection') \colon \theLanguageCollection' \subseteq \theLanguageCollection, |\theLanguageCollection'| = |\theLanguageCollection| - \theNumQueries\}$. Consider the following generator $\learner$. At the first timestep, while the consistent set contains two distinct languages $\theLanguage, \theLanguage'$ and queries remain, it queries a string $x \in \theLanguage \triangle \theLanguage'$, which is non-empty since the languages of $\theLanguageCollection$ are distinct. Exactly one of $\theLanguage, \theLanguage'$ contains $x$, so whatever the answer, one of the two becomes inconsistent with the target and is eliminated. Afterwards, the set $D$ of languages consistent with the answers has at most $|\theLanguageCollection| - \theNumQueries$ elements, so $D \subseteq \theLanguageCollection'$ for some $\theLanguageCollection' \subseteq \theLanguageCollection$ with $|\theLanguageCollection'| = |\theLanguageCollection| - \theNumQueries$. At every timestep, $\learner$ outputs an unseen string of the feedback-augmented closure whenever one exists.

By \cref{prop:unif-gen-feedback}, it suffices to show that for this $\learner$ there is no target, enumeration and timestep $\theTimestep$ with at least $\distinctNumExamples$ distinct strings at which the closure is finite, of size at least $\distinctNumExamples$, and contains only strings of $\theCurrentStringSet$. At any such timestep, the consistent set is a non-empty subset of $D$, since it contains the target, so the closure is a finite intersection of a non-empty subcollection of $\theLanguageCollection'$, and its size is below $B(\theLanguageCollection') \leq \distinctNumExamples$ by the definition of $B$. This contradicts its size being at least $\distinctNumExamples$, so the condition of \cref{prop:unif-gen-feedback} is never satisfied.
\end{proof}

\subsection{Uncomputable bound for context-free languages}

We showed that if the number of feedback queries is large enough, a collection can trivially become $0$-generatable with feedback. Nevertheless, there is no algorithm that computes a valid sample
bound for every finite collection of infinite context-free languages and every feedback budget.

\begin{restatable}{theorem}{CFLFeedbackGenThm}
    There is no algorithm that, given CFGs describing a finite collection $\theLanguageCollection$ of infinite context-free languages and a feedback budget $f$, always outputs $\distinctNumExamples$ such that $\theLanguageCollection$ is $\distinctNumExamples$-generatable with at most $f$ feedback queries.
\end{restatable}

\begin{proof}
Suppose, for contradiction, that such an algorithm exists. We show that it would decide the halting problem. We consider the feedback budget $\theNumQueries=1$.

Given a Turing machine $\mathcal{M}$, construct CFGs for languages $A$ and $B$ as in \cref{lemma:tm-construction}. In particular, if $\mathcal{M}$ halts in $s$ steps, then $|A\cap B|=2^s$. Since finiteness of context-free languages is decidable, we first check whether either language is finite.

If $A$ is finite, compute $|A|$ and choose $t\geq 1$ such that $2^t>|A|$. The machine cannot halt in $s\geq t$ steps, since this would imply $|A\cap B|=2^s\geq 2^t>|A|$, a contradiction. Therefore, simulating $\mathcal{M}$ for $t-1$ steps decides whether it halts. The case where $B$ is finite is analogous.

It remains to consider the case where both $A$ and $B$ are infinite. Let $\theAlphabet$ be their common alphabet and let $\$$ be a fresh symbol. We effectively construct CFGs for the following four languages, where each block $w_i$ belongs to $\theAlphabet^\ast$ and $w^R$ denotes
the reversal of $w$:
\begin{align*}
\theLanguage_1 &= \{w_1\$w_2\$w_3\$w_4 :
    w_1\in A,\ w_2^R\in B,\ w_3=w_4^R\},\\
\theLanguage_2 &= \{w_1\$w_2\$w_3\$w_4 :
    w_1=w_2^R,\ w_3\in A,\ w_4^R\in B\},\\
\theLanguage_3 &= \{w_1\$w_2\$w_3\$w_4 :
    w_1=w_4^R,\ w_2=w_3^R\},\\
\theLanguage_4 &= \{w_1\$w_2\$w_3\$w_4 :
    w_1\in B,\ w_2^R\in A,\ w_3\in B,\ w_4^R\in A\}.
\end{align*}
These languages are context-free. For $\theLanguage_1$ and $\theLanguage_2$, this follows from closure under reversal and concatenation, together with the
context-freeness of $\{u\$u^R:u\in\theAlphabet^\ast\}$.
The language $\theLanguage_3$ has nested reversal constraints and is recognized
by a pushdown automaton, while $\theLanguage_4$ is a concatenation of
context-free languages and separators. All four languages are infinite because $A$ and $B$ are infinite.

Apply the hypothetical algorithm to
$\theLanguageCollection=\{\theLanguage_1,\theLanguage_2,\theLanguage_3,\theLanguage_4\}$
and $\theNumQueries=1$, obtaining a bound $\distinctNumExamples$ such that
$\theLanguageCollection$ is $\distinctNumExamples$-generatable with one
feedback query. Choose $t\geq 1$ such that
$2^t\geq\distinctNumExamples$. We claim that $\mathcal{M}$ cannot
halt in $s\geq t$ steps.

Suppose otherwise, and write $H=A\cap B$, so that $|H|=2^s$. Every pairwise intersection of the $\theLanguage_i$ is finite: the combined constraints force each block to belong to $H$ or
$H^R$. For example, $\theLanguage_1\cap \theLanguage_2=\{x\$x^R\$y\$y^R:x,y\in H\}$ and $\theLanguage_1\cap \theLanguage_3
    =\{x\$x^R\$x\$x^R:x\in H\}$.
Moreover, the intersection of all four languages is $D=\{x\$x^R\$x\$x^R:x\in H\}$,
which has $2^s$ strings. Consequently, every subcollection of at least two languages has a finite intersection containing $D$.

Now fix any learner that asks at most one feedback query. We describe an adversary strategy against it. Since the learner is deterministic, this strategy determines a target $\theTrueLanguage\in\theLanguageCollection$
and an enumeration of $\theTrueLanguage$ on which the learner fails.

The adversary begins by enumerating $D$. Until the learner asks its query, all four languages remain consistent. When the learner asks its query, the adversary answers so that at least two languages stay consistent: for any queried string, at least two of the four languages contain it or at least two do not, and if both answers leave two languages, the adversary chooses either. This answer is consistent with every example already presented, since these are in all four languages. The adversary finishes enumerating $D$ and then enumerates the remaining strings in the intersection of the languages that remain consistent. Once this finite intersection is exhausted, the generator must output either a previously seen string or a string outside one of these
languages. In the first case it fails for any of them as the target, and in the second the adversary chooses a language that does not contain the output as the target $\theTrueLanguage$.

If the learner has not asked its query when $D$ is exhausted, it must either ask it before producing its output, in which case the preceding argument applies, or output without asking. Since the intersection of all four languages is exactly $D$, the latter also forces a
mistake for some target $\theTrueLanguage\in\theLanguageCollection$
consistent with the examples. In either case, the mistake occurs after at least $2^s$ distinct examples. All examples presented so far and any query answer are consistent with $\theTrueLanguage$, so the adversary can complete the enumeration by presenting its remaining strings and answering any subsequent query according to $\theTrueLanguage$.

Thus, $\theLanguageCollection$ is not
$\distinctNumExamples$-generatable with one feedback query, since $2^s\geq 2^t\geq\distinctNumExamples$. This contradicts the bound returned by the hypothetical algorithm. Therefore, $\mathcal{M}$ cannot halt in $s\geq t$ steps, and simulating it for $t-1$ steps decides whether it halts.

In all cases, we have decided the halting problem, a contradiction.
\end{proof}

\subsection{Bound for regular languages}

\begin{restatable}{theorem}{RegFeedbackGenThmUpperBound}
Let $\theLanguageCollection$ be a finite collection of infinite regular languages over an alphabet of size $\theAlphabetSize\geq 2$, such that each language $\theLanguage\in\theLanguageCollection$ is accepted by an NFA with at most $s$ states. Let $\theNumQueries<|\theLanguageCollection|-1$. Then $\theLanguageCollection$ is $\theAlphabetSize^{s^{|\theLanguageCollection|-\theNumQueries}}$-generatable with $\theNumQueries$ feedback queries.
\end{restatable}

\begin{proof}
By Theorem~4.1 of \citet{arenas2026languagegenerationcomplexitybarriers}, every collection $\theLanguageCollection' \subseteq \theLanguageCollection$ is $\theAlphabetSize^{s^{|\theLanguageCollection'|}}$-generatable, so $B(\theLanguageCollection') = \theAlphabetSize^{s^{|\theLanguageCollection| - \theNumQueries}}$ for every $\theLanguageCollection'$ with $|\theLanguageCollection'| = |\theLanguageCollection| - \theNumQueries$. The claim follows from \cref{thm:feedback-generic}.
\end{proof}

\subsection{Bounds for LTT languages}

\begin{restatable}{theorem}{LTTFeedbackGenThmUpperBound}
Let $\theLanguageCollection$ be a finite collection of infinite LTT languages over an alphabet of size $\theAlphabetSize\geq 2$, such that each language $\theLanguage\in\theLanguageCollection$ is defined by an LTT of size at most $s$. Let $\theNumQueries<|\theLanguageCollection|-1$. Then $\theLanguageCollection$ is $\theAlphabetSize^{3s(|\theLanguageCollection|-\theNumQueries)((s(|\theLanguageCollection|-\theNumQueries))^2+1)^2}$-generatable with $\theNumQueries$ feedback queries.
\end{restatable}

\begin{proof}
By Theorem~5.1 of \citet{arenas2026languagegenerationcomplexitybarriers}, every collection $\theLanguageCollection' \subseteq \theLanguageCollection$ of LTT languages of size at most $s$ is $\theAlphabetSize^{3r(r^2+1)^2}$-generatable with $r = s|\theLanguageCollection'|$, so $B(\theLanguageCollection') = \theAlphabetSize^{3r(r^2+1)^2}$ with $r = s(|\theLanguageCollection| - \theNumQueries)$ for every $\theLanguageCollection'$ with $|\theLanguageCollection'| = |\theLanguageCollection| - \theNumQueries$. The claim follows from \cref{thm:feedback-generic}.
\end{proof}

\LTTFeedbackGenThmLowerBound*

\begin{proof}
    Let $\theLanguageCollection$ be a collection of $n$ languages over the alphabet $\theAlphabet=\{1,\ldots, n\}$, where $\theLanguage_i=(\varepsilon,\varepsilon,\{(j, =1)\colon j\neq i\})$, i.e., it only leaves the symbol $i$ unconstrained and contains exactly one occurrence of the other symbols. Each language is infinite since the symbol $i$ can appear any number of times. 

    For any subset $\theLanguageCollSubset\subseteq \theLanguageCollection$ such that $|\theLanguageCollSubset|\geq 2$, the intersection $\bigcap_{\theLanguage\in\theLanguageCollSubset}\theLanguage$ forces every symbol that is constrained in at least one language to appear exactly once. Since for any two languages, every symbol is constrained in at least one of them, all $n$ symbols must appear exactly once. Therefore, the intersection is the set of all permutations of $\{1,\ldots,n\}$, which has size $n!$.

    No query eliminates more than one language: a string whose symbols all occur exactly once belongs to every $\theLanguage_i$, a string where exactly one symbol $i$ occurs a different number of times belongs to $\theLanguage_i$ only, and any other string belongs to none. In the first and third cases the answer is consistent with every remaining language, and in the second the adversary answers that the string is not in the target, eliminating only $\theLanguage_i$. Hence after $\theNumQueries<|\theLanguageCollection|-1$ queries at least two languages remain consistent, and by the previous paragraph the feedback-augmented closure is the set of permutations, of size $n!\geq 2^n$ for $n\geq 4$. Then $\theLanguageCollection$ is not $2^n$-generatable with $\theNumQueries$ feedback queries.
\end{proof}

\subsection{Bound for PO languages}

\begin{restatable}{theorem}{PTFeedbackGenThmUpperBound}
Let $\theLanguageCollection$ be a finite collection of infinite PO languages over an alphabet of size $\theAlphabetSize\geq 2$, such that each language $\theLanguage\in\theLanguageCollection$ is accepted by a PONFA with at most $s$ states. Let $\theNumQueries<|\theLanguageCollection|-1$. Then $\theLanguageCollection$ is $\theAlphabetSize^{(|\theLanguageCollection|-\theNumQueries)(s-1)+1}$-generatable with $\theNumQueries$ feedback queries.
\end{restatable}

\begin{proof}
By \cref{thm:lt-bound}, every collection $\theLanguageCollection' \subseteq \theLanguageCollection$ of PO languages, each recognized by a PONFA with at most $s$ states, is $\theAlphabetSize^{|\theLanguageCollection'|(s-1)+1}$-generatable, so $B(\theLanguageCollection') = \theAlphabetSize^{(|\theLanguageCollection| - \theNumQueries)(s-1)+1}$ for every $\theLanguageCollection'$ with $|\theLanguageCollection'| = |\theLanguageCollection| - \theNumQueries$. The claim follows from \cref{thm:feedback-generic}.
\end{proof}

\end{document}